\pdfoutput=1
\documentclass[11pt,letterpaper]{article}
\ifx\pdfminorversion\undefined\else\pdfminorversion=7\fi

\usepackage[letterpaper,left=1.05in,right=1.05in,top=1in,bottom=1in]{geometry}
\usepackage[T1]{fontenc}
\usepackage[utf8]{inputenc}
\usepackage[english]{babel}

\usepackage{mathpazo}
\usepackage{tgpagella}
\usepackage{inconsolata}
\usepackage{microtype}
\usepackage{amsmath,amsfonts,amssymb,mathtools}
\usepackage{graphicx}
\usepackage{booktabs}
\usepackage{multirow}
\usepackage{tabularx}
\usepackage{array}
\usepackage{wrapfig}
\usepackage{nicefrac}
\usepackage{xspace}
\usepackage[table,dvipsnames]{xcolor}
\usepackage{pifont}
\usepackage{fontawesome5}
\usepackage{tikz}
\usetikzlibrary{positioning,arrows.meta,fit,backgrounds,decorations.pathmorphing,calc}
\usepackage{algorithm}
\usepackage{algpseudocode}
\usepackage{tcolorbox}
\tcbuselibrary{skins,breakable}
\usepackage{enumitem}
\usepackage{placeins}
\usepackage{caption}
\usepackage{subcaption}
\usepackage{authblk}
\usepackage[explicit]{titlesec}
\usepackage{fancyhdr}
\usepackage{lastpage}

\PassOptionsToPackage{authoryear,round}{natbib}
\usepackage{natbib}

\usepackage[bookmarks=true,bookmarksopen=true]{hyperref}

\definecolor{inkblack}  {HTML}{0A0A0A}   
\definecolor{inkdark}   {HTML}{1A1A1A}   
\definecolor{inkmid}    {HTML}{2B2B2B}   
\definecolor{inkgrey}   {HTML}{4A4A4A}   
\definecolor{inkmuted}  {HTML}{6E6E6E}   
\definecolor{inkfaint}  {HTML}{8C8C8C}   
\definecolor{inkrule}   {HTML}{BFBFBF}   
\definecolor{inkbg}     {HTML}{F7F7F7}   
\definecolor{inkbg2}    {HTML}{F0F0F0}   
\definecolor{inkframe}  {HTML}{D9D9D9}   

\colorlet{narrared}{inkmid}
\colorlet{narrareddeep}{inkblack}
\colorlet{cforge}{inkdark}
\colorlet{cmemory}{inkgrey}
\colorlet{cpace}{inkmid}
\colorlet{cmeta}{inkmuted}
\colorlet{cnovelty}{inkdark}
\definecolor{ctablehead}{HTML}{D6D6D6}
\definecolor{ctablerule}{HTML}{BFBFBF}
\colorlet{ctabletext}{inkdark}
\colorlet{cstoryq}{inkblack}
\colorlet{cux}{inkgrey}
\colorlet{openyes}{inkblack}
\colorlet{closedno}{inkfaint}
\definecolor{HeaderGray}{RGB}{238,238,238}
\definecolor{RowGray}{RGB}{242,244,248}
\definecolor{AnchorBlue}{RGB}{240,247,255}
\definecolor{SiblingGreen}{RGB}{244,250,246}
\definecolor{CircleBlue}{RGB}{41,98,168}
\definecolor{CircleOrange}{RGB}{204,102,0}
\definecolor{CircleTeal}{RGB}{0,128,128}

\hypersetup{
    colorlinks  = true,
    linkcolor   = inkblack,
    citecolor   = inkmid,
    urlcolor    = inkmid,
    filecolor   = inkblack,
    pdfborder   = {0 0 0}
}

\titleformat{\section}
  {\Large\bfseries\color{inkblack}}{\thesection}{0.6em}{#1}[]
\titleformat{\subsection}
  {\large\bfseries\color{inkmid}}{\thesubsection}{0.55em}{#1}[]
\titleformat{\subsubsection}
  {\normalsize\bfseries\itshape\color{inkmid}}{\thesubsubsection}{0.5em}{#1}[]
\titlespacing*{\section}      {0pt}{2.6ex plus 0.8ex minus 0.4ex}{1.0ex}
\titlespacing*{\subsection}   {0pt}{2.0ex plus 0.7ex minus 0.3ex}{0.8ex}
\titlespacing*{\subsubsection}{0pt}{1.5ex plus 0.5ex minus 0.2ex}{0.6ex}

\setlist[itemize,1]{label=\textbullet,leftmargin=1.4em,topsep=2pt,itemsep=2pt,parsep=0pt}
\setlist[itemize,2]{label=$\circ$,leftmargin=1.2em}
\setlist[enumerate,1]{leftmargin=1.6em,topsep=2pt,itemsep=2pt,parsep=0pt}

\renewcommand\Affilfont{\small\color{inkmuted}}
\makeatletter
\renewcommand\AB@affilsepx{,\enspace\protect\Affilfont}
\makeatother

\fancypagestyle{firststyle}{%
  \fancyhf{}%
  \fancyfoot[C]{\footnotesize\color{inkmuted}\thepage\ /\ \pageref{LastPage}}%
}

\newcommand{\site}[1]{\texttt{#1}}
\newcommand{\UXBench}{\textsc{UXBench}\xspace}

\newcommand{\cmark}[1]{%
  \tikz[baseline=(char.base)]{%
    \node[shape=circle, fill=CircleBlue!15, draw=CircleBlue,
          inner sep=1.6pt, font=\scriptsize\bfseries, text=CircleBlue] (char) {#1};}%
\xspace}
\newcommand{\omark}[1]{%
  \tikz[baseline=(char.base)]{%
    \node[shape=circle, fill=CircleOrange!12, draw=CircleOrange,
          inner sep=1.6pt, font=\scriptsize\bfseries, text=CircleOrange] (char) {#1};}%
\xspace}
\newcommand{\tmark}[1]{%
  \tikz[baseline=(char.base)]{%
    \node[shape=circle, fill=CircleTeal!12, draw=CircleTeal,
          inner sep=1.6pt, font=\scriptsize\bfseries, text=CircleTeal] (char) {#1};}%
\xspace}

\newcommand{\cnum}[1]{\ding{\the\numexpr181+#1\relax}}

\title{\textsc{UXBench}: Measuring the Actionability of LLM-Generated UX Critiques}

\author[1]{Wenjie Wang}
\author[1]{Yue Huang}
\author[2]{Zipeng Ling}
\author[1]{Han Bao}
\author[3]{Hang hua}
\author[1]{Xiaonan Luo}
\author[1]{Yu Jiang}
\author[4]{Shiyi Du}
\author[5]{Yuexing Hao}
\author[6]{Xiaomin Li}
\author[7]{Yuchen Ma}
\author[6]{Dianzhuo Wang}
\author[1]{Yanfang Ye}
\author[1]{Xiangliang Zhang\textsuperscript{*}}

\affil[1]{University of Notre Dame}
\affil[2]{University of Pennsylvania}
\affil[3]{University of Rochester}
\affil[4]{Carnegie Mellon University}
\affil[5]{Massachusetts Institute of Technology}
\affil[6]{Harvard University}
\affil[7]{LMU Munich}

\newcommand{\equalcontributionnote}{%
  \textsuperscript{*}\,Corresponding author: \texttt{xzhang33@nd.edu}.\\[2pt]}

\newcommand{\paperurl}{https://github.com/Jackwwj619/UXBench}
\newcommand{\paperurllabel}{Code}
\newcommand{\projecturl}{https://jackwwj619.github.io/projects/UXBench/}
\newcommand{\projecturllabel}{Project Page}

\makeatletter
\newcommand{\makecoverpage}{%
  \thispagestyle{firststyle}%
  \vspace*{-1.0em}%

  {\noindent\bfseries\color{inkblack}\fontsize{18pt}{23pt}\selectfont
   \@title\par}
  \vspace{1.0em}

  {\noindent\@author\par}
  \vspace{0.35em}
  \ifx\equalcontributionnote\@empty\else
    {\noindent\footnotesize\itshape\color{inkmuted}\equalcontributionnote\par}%
    \vspace{0.15em}%
  \fi
  {\centering\color{inkmid}%
   \faGithub\ \href{\paperurl}{\paperurllabel}\hspace{1.4em}%
   \faGlobe\ \href{\projecturl}{\projecturllabel}\par}
  \vspace{1.2em}

  \begin{tcolorbox}[
      enhanced, sharp corners, width=\textwidth,
      colback=inkbg, colframe=inkbg, boxrule=0pt,
      left=16pt,right=16pt,top=12pt,bottom=12pt
  ]
    {\bfseries\color{inkblack}\large Abstract}\par
    \vspace{0.45em}
    {\small\color{inkdark}\theabstract\par}
  \end{tcolorbox}
  \vspace{0.8em}
}
\makeatother

\newcommand{\theabstract}{%
Large language models (LLMs) are increasingly deployed as UX judges that inspect interfaces, diagnose usability problems, and propose repairs. Yet no controlled benchmark measures whether the resulting critiques are reliable and actionable across heterogeneous product surfaces. We introduce \UXBench, a benchmark for evaluating LLMs as interaction-grounded UX judges. \UXBench comprises local-first runnable web fixtures spanning ten product-surface families, paired with coverage-gated browser exploration that forces models to collect interaction evidence before reporting. Each judge model produces a structured UX report over seven rubric dimensions; report quality is measured by whether a fixed downstream repair agent can improve the interface based on the critique. We evaluate eight frontier models under both an automated repair-lift protocol and a blind human validation study. Results show that UX judging is neither saturated nor one-dimensional: models differ meaningfully in report actionability, exhibit distinct rubric-level repair signatures, vary in fixture-level reliability, and trade leadership across surface categories.
}

\begin{document}

\makecoverpage

\section{Introduction}

\begin{wrapfigure}{t}{0.5\linewidth}
    \centering
    \includegraphics[width=\linewidth]{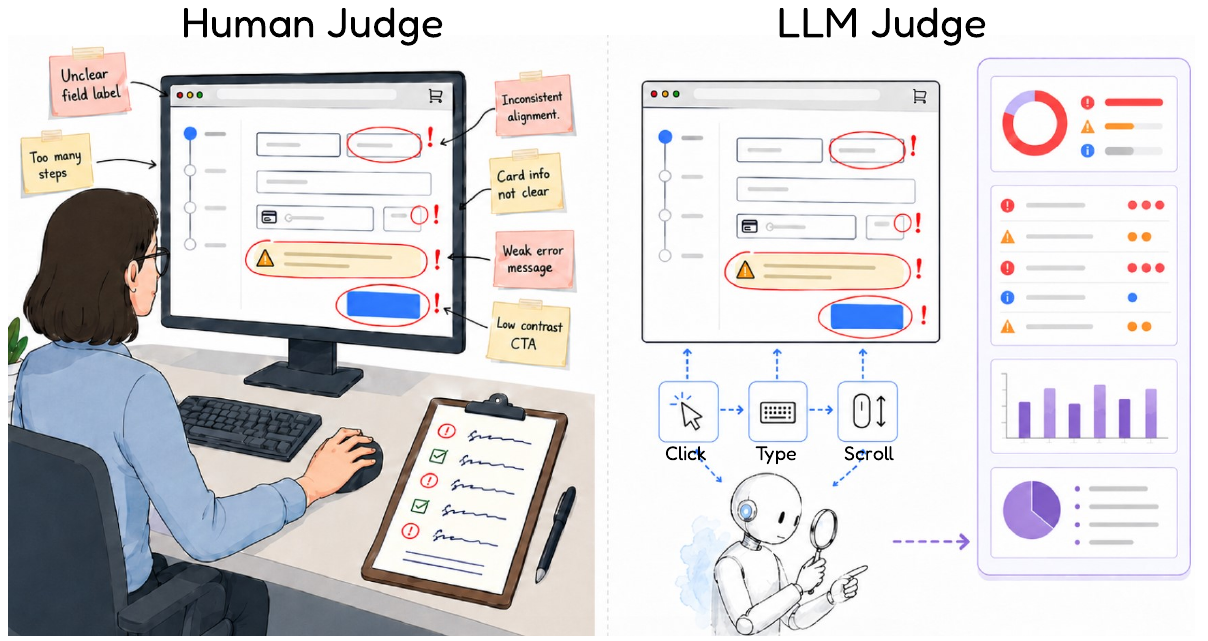}
    \caption{Interaction-grounded UX judging: like human reviewers, LLM judges should inspect the running interface, exercise controls, and ground critiques in observed feedback rather than static impressions.}
    \label{fig:intro}
    \vspace{-10pt}
\end{wrapfigure}

LLMs are increasingly used not only to generate code and interface mockups but also to review existing interfaces, diagnose usability problems, and propose repairs~\citep{beltramelli2017pix2code, si2025design2code, wu2024uicoder, duan2024uicrit}. This shift makes UX judgment a natural target for model evaluation~\citep{liu2023geval, zheng2023judging}. Recent work further extends this direction from static UI generation and critique to LLM-assisted usability testing, simulated user studies, and automated UX flaw detection~\citep{lu2025uxagent, calvano2025leverage, gao2026trainingcomputeruseagents}. A useful judge should browse a page, recognize when users cannot infer the current state or next action, identify missing feedback or recovery paths, and communicate these issues in a form developers can act on~\citep{parkla2021evaluating, shinei1983direct}. Yet it remains unclear whether current frontier models are reliable UX judges, or whether models that appear similar in general capability produce equally actionable critiques. As illustrated in \autoref{fig:intro}, we study UX judging as an interaction-grounded evaluation problem: both human and LLM judges must inspect the running interface, exercise controls, observe feedback, and turn the resulting evidence into actionable critique.

Evaluating this capability is difficult because web UX is not fully visible from a screenshot or a static DOM snapshot. Many usability failures are interactional: a disabled control may have no explanation, a form may silently reject input, a destructive action may hide its consequences, or a mobile layout may collapse into an unusable state. Recent studies of digital forms and responsive interfaces show that usability depends on input organization, validation behavior, editing flow, and cross-device layout behavior, rather than only on the first rendered view~\citep{Iftikhar2020Comparing, breitmayer2024exploring, parkla2021evaluating}. Accessibility work similarly treats labels, error identification, focus behavior, status messages, and dynamic content changes as properties that must remain perceivable during interaction~\citep{w3c2024wcag22, mehralian2025dynamic}. A judge that only inspects the first rendered view can therefore produce fluent but weakly grounded criticism. Reliable UX judging requires evidence collection: the model must exercise controls, observe state transitions, test error and recovery paths, and tie each finding to concrete behavior in the running interface.

Existing benchmarks only partially cover this setting. Web and GUI agent benchmarks typically measure whether an agent can complete user tasks in a browser or operating-system environment~\citep{deng2023mind2web, zhou2024webarena, koh2024visualwebarena, he2024webvoyager, xie2024osworld, cheng2024seeclick}, while UI generation benchmarks evaluate whether models can synthesize interfaces from textual or visual specifications~\citep{beltramelli2017pix2code, si2025design2code, yun2024web2code}. Screenshot-level preference and design-evaluation setups can compare visual appeal or layout quality, or region-level design critiques~\citep{wu2024uiclip, duan2024uicrit}, but they do not isolate whether a model can produce interaction-grounded UX reports whose recommendations lead to better interfaces. As a result, the community lacks a controlled benchmark for measuring the actionability and reliability of LLM-generated UX critique across the heterogeneous product surfaces that users encounter in practice.

We introduce \UXBench, a benchmark for evaluating LLMs as interaction-grounded UX judges. \UXBench consists of local-first runnable web fixtures implemented as static HTML/CSS/JavaScript bundles, removing live-site drift from A/B tests, personalization, backend failures, and third-party outages. The benchmark spans ten surface families---from landing pages and checkout flows to dashboards, chatbot interfaces, and mobile micro-UIs. Within each family, \UXBench pairs real-product anchors with independently authored synthetic siblings: anchors preserve recognizable interaction patterns, while siblings vary branding, text, layout, and visual identity so that models must judge the interface in front of them rather than recall memorized impressions of familiar products.

Each \UXBench run evaluates a judge model through a coverage-gated browser trajectory. The model prescans the fixture, forms an exploration plan, then repeatedly observes the rendered page, takes user-like actions, and inspects the resulting feedback. A coverage gate prevents premature termination: if the model attempts to stop before collecting sufficient evidence, the environment returns the unmet conditions and resumes exploration. After the trajectory, the model produces an evidence-grounded UX report over seven rubric dimensions. To measure whether the critique is actionable rather than merely plausible, \UXBench passes the report to a fixed repair agent that edits the fixture while preserving the original product intent, brand identity, and interaction semantics. The repaired interface is then scored under a fixed evaluator, turning report actionability into a comparable signal.

We evaluate eight frontier models with two complementary protocols: an automated sweep and a blind human validation study. Across both, \UXBench reveals that UX judging is neither saturated nor one-dimensional: repair lift varies meaningfully across models, rubric-level signatures diverge, fixture-level reliability differs even among similarly ranked models, and the leading model changes across product-surface categories.

Our contributions are threefold:

\vspace{2pt}
\noindent\cmark{1} We formulate LLM-based UX judging as an interaction-grounded, report-conditioned repair evaluation problem, where the quality of a judge is measured by whether its evidence-backed critique supports downstream interface improvement.

\vspace{2pt}
\noindent\cmark{2} We introduce \UXBench, a local-first suite of runnable web fixtures spanning ten product-surface families, with real anchors, synthetic siblings, coverage-gated browser exploration, and evidence-grounded UX reporting.

\vspace{2pt}
\noindent\cmark{3} We evaluate eight frontier models under automated and human protocols, showing that UX judge models differ in report actionability, rubric-level strengths, fixture-level reliability, and surface-conditioned competence.

\section{Related Work}

\subsection{Web and GUI Agent Benchmarks}

Recent web and GUI agent benchmarks evaluate whether language and multimodal agents can perceive interface state, act in interactive environments, and complete user-specified tasks. Early controlled platforms such as World of Bits and MiniWoB/MiniWoB++ established web interaction as a measurable learning problem, while later benchmarks extend the setting to realistic websites, visual web environments, enterprise software, desktop operating systems, and mobile devices. Collectively, Mind2Web, WebArena, VisualWebArena, WebVoyager, SeeClick, OSWorld, WorkArena, BrowserGym, WebLINX, Android in the Wild, AndroidWorld, and GUIOdyssey measure abilities such as instruction following, multi-step navigation, visual grounding, desktop control, conversational web interaction, and cross-app mobile operation~\citep{shi2017worldofbits, liu2018reinforcementweb, deng2023mind2web, zhou2024webarena, koh2024visualwebarena, he2024webvoyager, cheng2024seeclick, xie2024osworld, drouin2024workarena, dechezelles2024browsergym, lu2024weblinx, rawles2025androidworlddynamicbenchmarkingenvironment, rawles2024androidwild, lu2025guiodyssey}. These benchmarks make interaction a central evaluation object, but their success criteria are usually task completion, final-state correctness, or action accuracy. In contrast, \UXBench studies a different capability: the model is not asked to complete a user task, but to inspect a running interface, gather evidence about usability failures, and produce findings that can support downstream interface repair.

\subsection{UI Generation and Automated Design}

Another line of work studies interface understanding, UI generation, and automated design evaluation. Large-scale UI datasets and design-search systems such as Rico, learned mobile design semantics, WebUI, Gallery D.C., and VINS support component understanding, semantic annotation, retrieval, visual search, and data-driven design assistance~\citep{deka2017rico, liu2018designsemantics, wu2023webui, chen2019gallerydc, bunian2021vins}. UI generation methods and benchmarks, including pix2code, Design2Code, Web2Code, UICoder, and DCGen, further evaluate whether models can synthesize interface code from screenshots, webpages, text specifications, or multimodal prompts~\citep{beltramelli2017pix2code, si2025design2code, yun2024web2code, wu2024uicoder, wan2024dcgen}. Closer to evaluation, UIClip and UICrit assess visual design quality, screenshot--description alignment, or region-level design critiques~\citep{wu2024uiclip, duan2024uicrit}. This work shows that models can reason over UI structure, appearance, and code, but the evaluation target is often the generated artifact or a static visual/design judgment. In contrast, \UXBench evaluates critiques produced after browser interaction: missing feedback, silent validation, hidden consequences, weak recovery paths, and other UX failures are treated as interactional evidence that should be observed, reported, and tested through report-conditioned repair.

\subsection{LLM-Assisted Usability Testing}

Recent work has begun to use LLMs and multimodal agents for usability testing, heuristic evaluation, accessibility testing, and simulated user studies. Earlier automated-usability-evaluation research surveyed how software tools can support usability inspection, testing, inquiry, analytical modeling, and simulation, while also emphasizing the difficulty of automating context-sensitive interface judgment~\citep{ivory2001state}. Newer LLM-based systems revisit this problem with stronger language and UI-understanding capabilities: UXAgent simulates usability testing with LLM agents, AXNav replays natural-language accessibility tests through assistive-technology navigation, and recent studies explore LLM-based UX testing, heuristic evaluation, and comparison with human experts~\citep{lu2025uxagent, taeb2024axnav, hsueh2024applyingllm, guerino2025gpt4ousability}. These works are closest to \UXBench because they treat usability evaluation as an interactive or expert-like process rather than a purely visual judgment. \UXBench differs mainly in its evaluation framing: instead of using LLMs as open-ended testing assistants, it fixes the fixtures, exploration protocol, repair agent, and scorer to compare judge models by whether their evidence-grounded reports produce measurable interface improvement.

\section{Benchmark Construction}

\begin{figure*}[t]
  \centering
  \includegraphics[width=\textwidth]{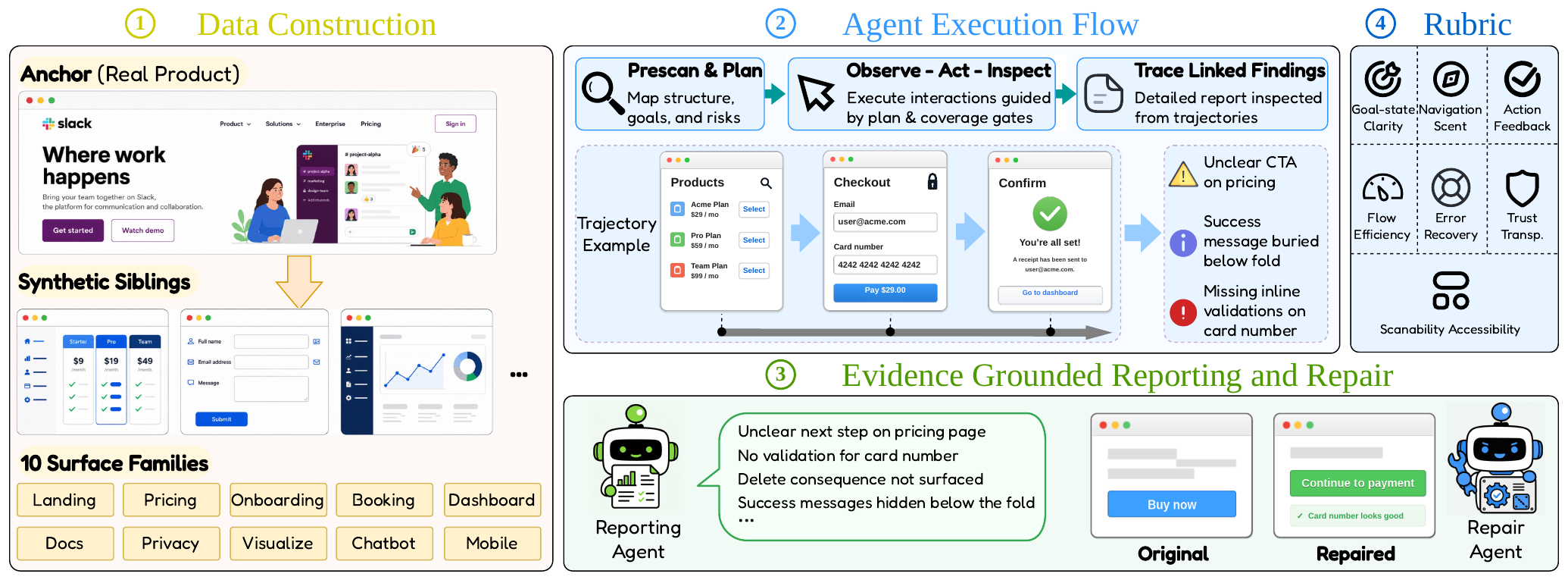}
  \caption{\label{fig:method}
  Overview of \UXBench. Local-first real anchors and synthetic siblings are evaluated through coverage-gated browser exploration. Trace-linked findings are converted into evidence-grounded UX reports, which are then used for report-conditioned repair and scored with the same seven-dimensional UX rubric.}
  \vspace{-10pt}
\end{figure*}

\autoref{fig:method} provides an overview of the \UXBench construction and evaluation pipeline. The benchmark is built around four connected components: local-first fixture construction, coverage-gated browser exploration, evidence-grounded UX reporting, and report-conditioned repair.

\subsection{Fixture Design and Catalog}

\noindent \textbf{Local-First Fixtures.}
\UXBench consists of local-first web fixtures implemented as static HTML/CSS/JavaScript bundles, each servable from a local file server without backend dependencies. Pinning the interface in this way removes the major sources of drift that confound live-site evaluation (e.g., A/B tests, personalization, and third-party outages), so that any score difference between models reflects the models themselves rather than the environment in which they were evaluated.

\noindent \textbf{Surface Categories.}
Rather than building one or two deep applications, \UXBench prioritizes breadth across the kinds of product surfaces web users routinely encounter (e.g., marketing, transactional, operational, and mobile surfaces; the full taxonomy is given in \autoref{tab:catalog}). Different surfaces stress qualitatively different UX competences: a judge that handles one family well (e.g., a static marketing page) may still misread another (e.g., a transactional flow whose failures only emerge mid-interaction).

\noindent \textbf{Anchors and Synthetic Siblings.}
Within each surface category, \UXBench pairs one or more real-product anchors with several independently authored synthetic siblings (\autoref{tab:catalog}). The anchor grounds the category in a recognizable interaction pattern (e.g., the structure of a familiar checkout or dashboard), while the siblings preserve that interaction model but vary the surface (e.g., re-skinning the visual identity and rewriting the textual content). The anchor supplies external validity, while the siblings test whether the judge evaluates the interface currently in front of it rather than memorized impressions of well-known products. More details are provided in \autoref{app:surfaceCategory}.

\subsection{Agent Execution Flow}
\label{ssec:flow}

\noindent \textbf{Overview.}
A single \UXBench run proceeds in four stages: \tmark{1}~the judge agent executes a lightweight \emph{prescan} and produces an exploration \emph{plan}; \tmark{2}~it enters a closed-loop \emph{observe--act trajectory} in a real browser; \tmark{3}~on termination it emits an evidence-grounded UX \emph{report}; and \tmark{4}~the report is consumed by a fixed code-editing agent to perform localized \emph{repair} on the fixture source.

\noindent \textbf{Prescan and Planning.}
Each run begins with a prescan that compiles a brief structural summary of the site (e.g., visible pages and salient controls) and converts it into an explicit exploration plan stating what the agent intends to verify and when it may stop. The plan is adaptive, not scripted: the agent re-routes when feedback is ambiguous (e.g., switching to a recovery path or revisiting under a mobile viewport) rather than following a fixed click sequence.

\noindent \textbf{Browser-Based Trajectory.}
Following the prescan and plan, the agent evaluates the runnable interface by repeatedly observing the current page state, taking a user-like action, and inspecting the resulting feedback. Interactive evaluation is necessary because many UX failures only emerge during interaction (e.g., silent form validation or a disabled control with no visible response). Formally, at step $t$ the agent receives an observation $o_t$ of the current page (combining textual, structural, and visual cues, e.g., rendered text, a screenshot, and any runtime warnings) and selects an action $a_t$ (e.g., a click, a text entry, or a viewport switch), yielding an interaction trajectory
\[
\tau=(o_0,a_0,o_1,\ldots,o_T).
\]
\UXBench treats this trajectory as the primary evidence source: reported issues are expected to correspond to concrete interaction events (e.g., a state transition that occurred without feedback) rather than to unsupported visual impressions.

\begin{algorithm}[t]
\caption{Coverage-Gated Exploration}
\label{alg:coverage_gate}
\small
\begin{algorithmic}[1]
\Require fixture profile $P$ (HTML pages, prescan controls), plan $\Pi$ (phases, target pages), step budget $B$, depth $d{\in}\{\textsc{quick},\textsc{std},\textsc{deep},\textsc{exh}\}$, viewport mode $v{\in}\{\textsc{desk},\textsc{mob},\textsc{both}\}$
\State open target URL; $o_0\gets\textsc{Observe}()$; $\tau\gets[o_0]$
\For{$t=0,\ldots,B-1$}
    \State $C\gets\textsc{CoverageState}(P,\Pi,\tau,o_t,d,v)$
    \State $a_t\gets\textsc{Brain}(o_t,\tau,\Pi,C)$ \Comment{judge model proposes action}
    \If{$a_t=\textsc{finish}$ \textbf{and} $C.\textit{stop\_allowed}{=}\bot$} \Comment{retry with unmet coverage}
        \State $a_t\gets\textsc{Brain}(o_t,\tau,\Pi,C,\,\textit{must\_continue}{=}C.\textit{reason})$
    \EndIf
    \If{$a_t=\textsc{finish}$ \textbf{and} $C.\textit{stop\_allowed}{=}\bot$} \Comment{deterministic continuation}
        \State $a_t\gets\textsc{Fallback}(C)$: open an unvisited page, else exercise an unexplored control
    \EndIf
    \If{$a_t=\textsc{finish}$} \textbf{break} \EndIf
    \State execute $a_t$; $o_{t+1}\gets\textsc{Observe}()$
    \State append $a_t$ and $o_{t+1}$ to $\tau$
\EndFor
\State $C\gets\textsc{CoverageState}(P,\Pi,\tau,\textsc{LastObs}(\tau),d,v)$
\State \Return $\tau$,\, $\textsc{EvidenceConfidence}(\tau,C)\in\{\text{low},\text{med},\text{high}\}$
\end{algorithmic}
\end{algorithm}

\subsection{Evidence-Grounded Reporting and Repair}
\label{ssec:report}

After exploration, \UXBench produces a final UX report using the accumulated trajectory, coverage summary, and final browser state. Each metric-level judgment must be supported by concrete trace evidence, so a finding is considered valid only when it can be linked to an observed interaction event. This makes the report auditable rather than merely rhetorical. To further test whether the critique is actionable, \UXBench performs report-conditioned repair. Given the ranked findings and the fixture source code, we invoke Claude Code as an independent code-editing agent with a restricted file-editing toolset. The original interface, brand identity, and interaction intent are specified as invariants, ensuring that any improvement in repaired UX quality reflects the actionability of the generated report rather than an unconstrained redesign.

\subsection{Coverage-Gated Termination}
\label{ssec:coverage}

The trajectory loop terminates only when the collected evidence is substantive enough to back the report. \UXBench enforces this with a coverage gate: at each step, the system recomputes a compact coverage state that operationalizes evidential sufficiency over the interaction trace and the current plan (e.g., whether the agent has actually exercised the salient controls of the page, not merely visited it). If the judge proposes to stop before this threshold is reached, the environment rejects the \textsc{finish} decision and returns the unmet conditions, pushing the agent back toward under-explored regions. The gate therefore constrains only evidence collection, not the downstream report or repair stage, and its role is to ensure that final UX findings are grounded in observed interaction consequences rather than in superficial browsing. Algorithm~\ref{alg:coverage_gate} summarizes this loop; detailed algorithm is provided in \autoref{app:algoDetail}.

\definecolor{HeaderBlue}{RGB}{43,64,92}
\definecolor{RowGray}{RGB}{242,244,248}

\newcommand{\logosize}{0.95em}
\newcommand{\modelcell}[2]{%
  \raisebox{-0.16em}{\includegraphics[height=\logosize]{figure/logo/#2}}%
  \hspace{0.25em}\texttt{#1}%
}

\begin{table*}[t]
  \centering
  \small
  \setlength{\tabcolsep}{3.6pt}
  \renewcommand{\arraystretch}{1.32}
  \begin{tabularx}{\textwidth}{
    @{}
    >{\raggedright\arraybackslash}p{0.145\textwidth}
    >{\centering\arraybackslash}X
    >{\centering\arraybackslash}X
    >{\centering\arraybackslash}X
    >{\centering\arraybackslash}X
    >{\centering\arraybackslash}X
    >{\centering\arraybackslash}X
    >{\centering\arraybackslash}X
    >{\centering\arraybackslash}X
    >{\centering\arraybackslash}X
    >{\centering\arraybackslash}X
    @{}
  }
    \toprule
    & \multicolumn{3}{c}{\textbf{\textcolor{HeaderBlue}{Overall}}}
    & \multicolumn{7}{c}{\textbf{\textcolor{HeaderBlue}{Per-dimension ($\Delta$)}}} \\
    \cmidrule(lr){2-4}\cmidrule(lr){5-11}

    \textbf{Model}
    & \textbf{Base.}
    & \textbf{Repaired}
    & \textbf{$\Delta$}
    & \textbf{Goal}
    & \textbf{Nav}
    & \textbf{Fdbk}
    & \textbf{Flow}
    & \textbf{Err}
    & \textbf{Trust}
    & \textbf{Scan} \\
    \midrule

    \modelcell{GPT 5.4}{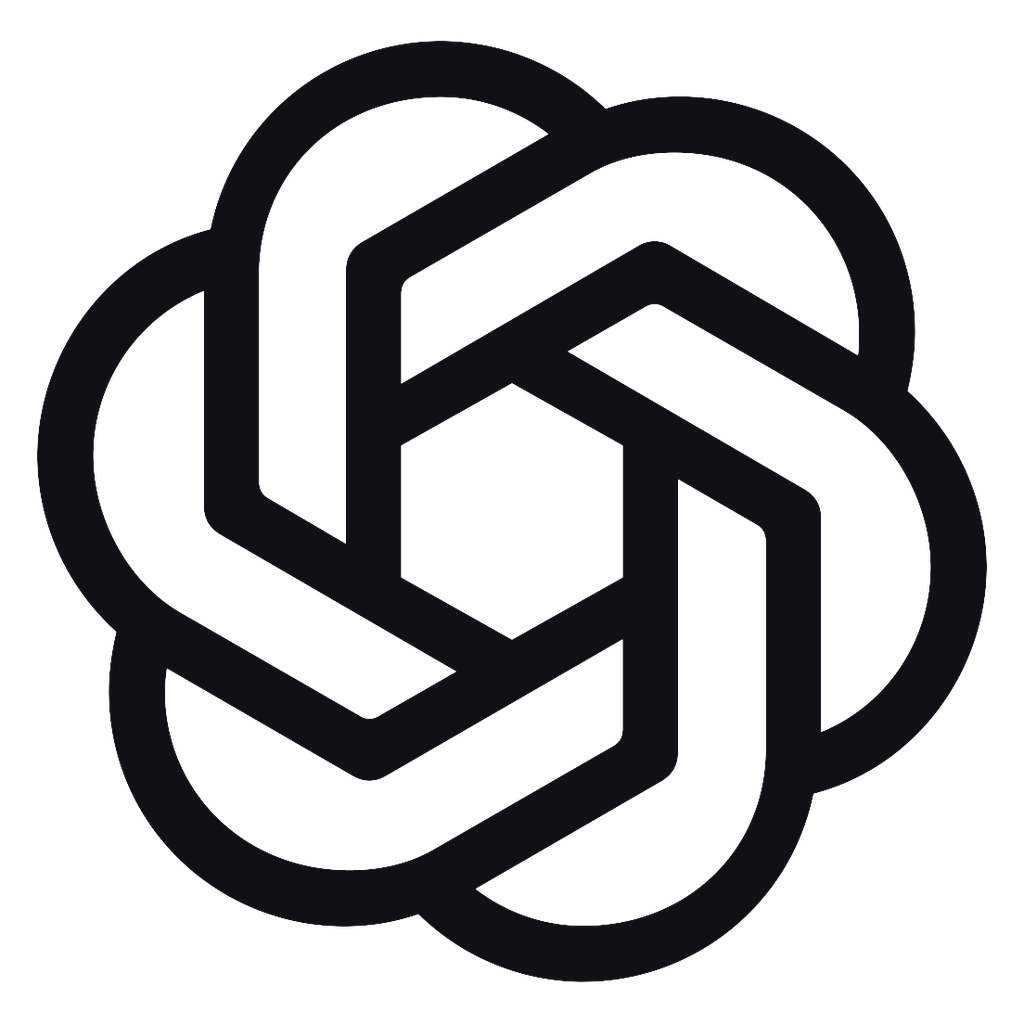}
    & 3.27 & 3.49 & \textbf{+0.22}
    & +0.13 & +0.15 & +0.36 & +0.05 & \textbf{+0.36} & +0.21 & +0.26 \\

    \rowcolor{RowGray}
    \modelcell{Kimi k2.5}{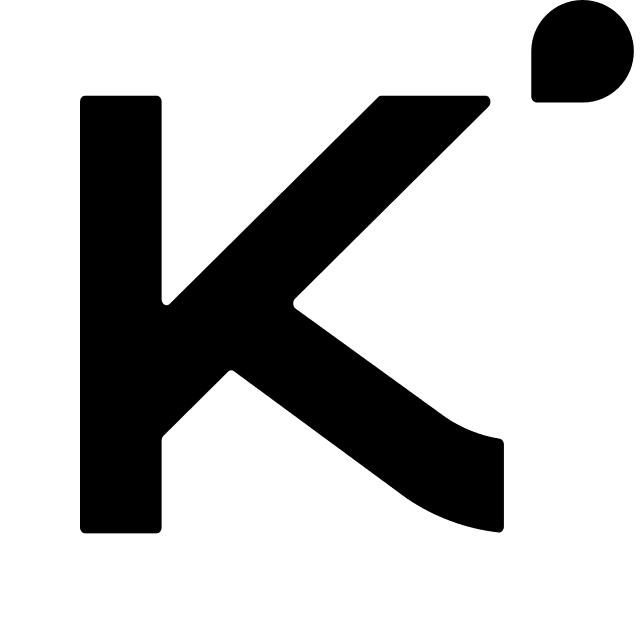}
    & 3.27 & 3.48 & +0.21
    & +0.12 & \textbf{+0.17} & \textbf{+0.44} & +0.05 & +0.12 & \textbf{+0.27} & +0.29 \\

    \modelcell{Sonnet 4.6}{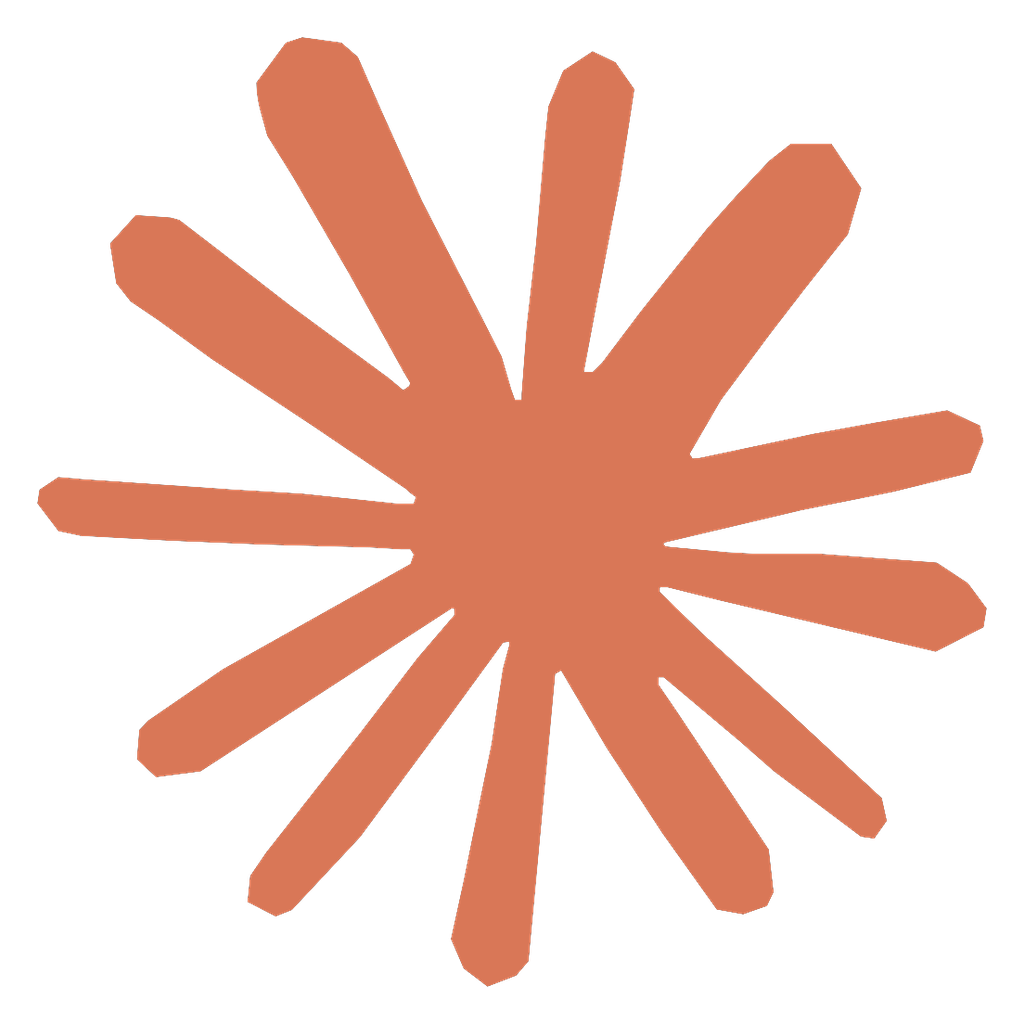}
    & 3.27 & 3.45 & +0.19
    & \textbf{+0.15} & +0.05 & +0.35 & +0.03 & +0.28 & +0.15 & +0.35 \\

    \rowcolor{RowGray}
    \modelcell{GPT Mini}{openai}
    & 3.27 & 3.45 & +0.19
    & +0.13 & +0.13 & +0.38 & +0.05 & +0.18 & +0.13 & +0.33 \\

    \modelcell{GLM 5.1}{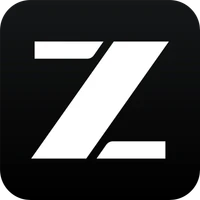}
    & 3.27 & 3.45 & +0.17
    & $-0.02$ & \textbf{+0.17} & +0.32 & +0.05 & +0.20 & +0.20 & +0.32 \\

    \rowcolor{RowGray}
    \modelcell{GPT Nano}{openai}
    & 3.27 & 3.44 & +0.17
    & +0.07 & +0.05 & +0.34 & +0.10 & +0.20 & +0.05 & +0.37 \\

    \modelcell{Qwen 3.6}{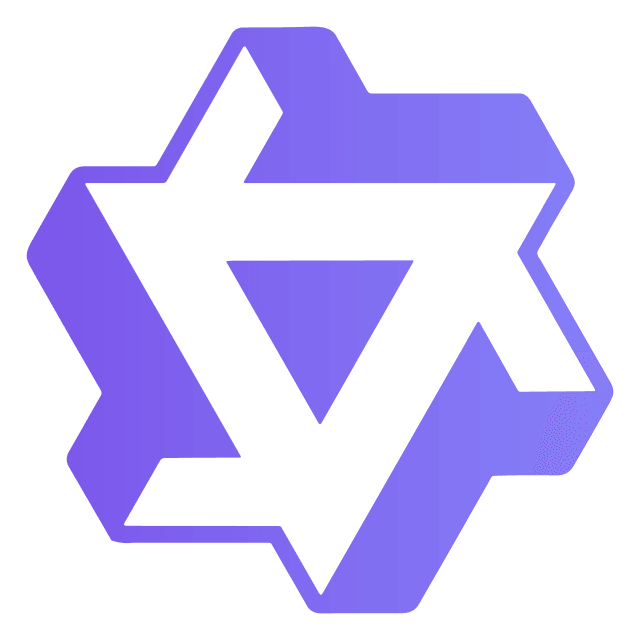}
    & 3.27 & 3.42 & +0.15
    & +0.02 & +0.07 & +0.24 & \textbf{+0.12} & +0.07 & +0.12 & \textbf{+0.39} \\

    \rowcolor{RowGray}
    \modelcell{Gemini 3.1}{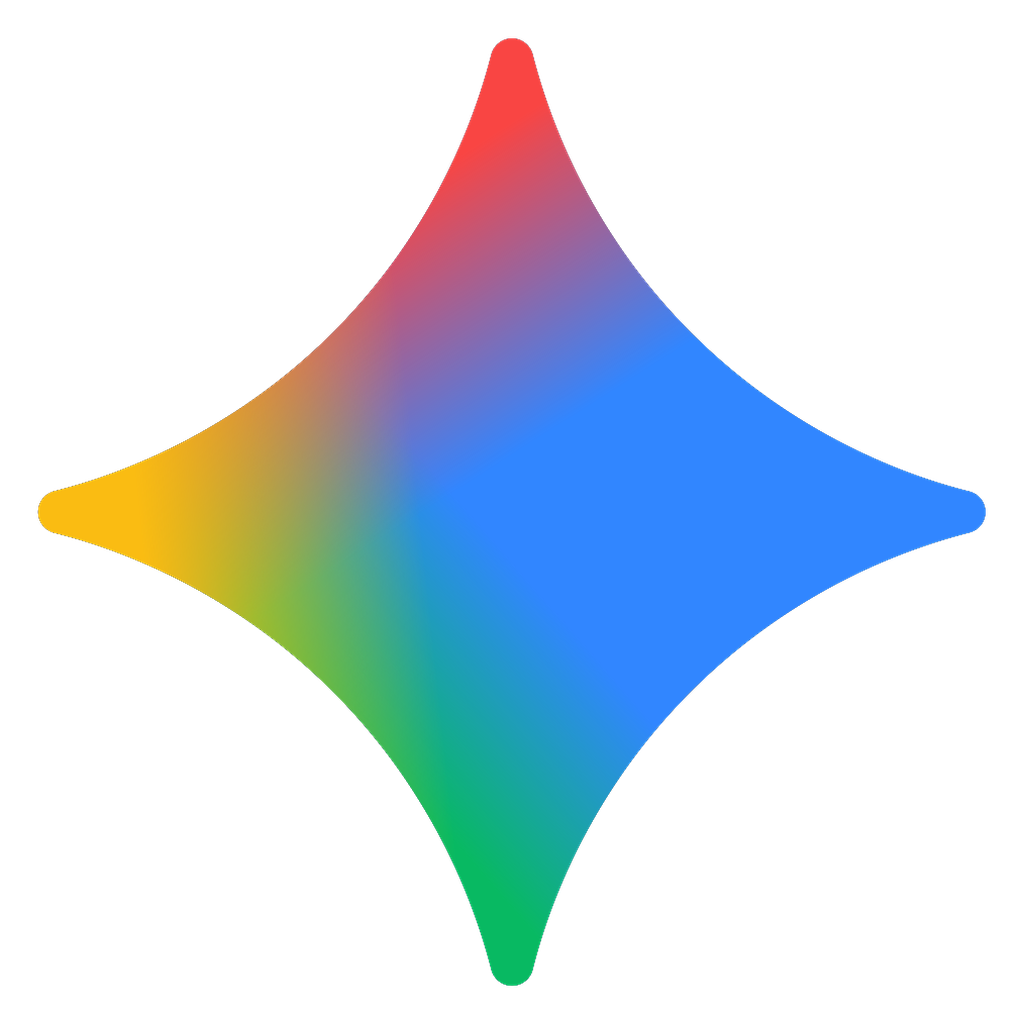}
    & 3.27 & 3.41 & {+0.14}
    & +0.08 & +0.03 & +0.24 & +0.08 & +0.11 & +0.11 & +0.34 \\

    \bottomrule
  \end{tabularx}
  \caption{\label{tab:overall}
  Overall and per-dimension repair lift across the eight evaluated models, sorted by $\Delta$. \textbf{Bold} indicates the column maximum. For compactness, tables abbreviate Claude-Sonnet-4.6 as \texttt{Sonnet 4.6}, GPT-5.4-Mini as \texttt{GPT Mini}, GPT-5.4-Nano as \texttt{GPT Nano}, Qwen-3.6-Plus as \texttt{Qwen 3.6}, and Gemini-3.1-Pro as \texttt{Gemini 3.1}.}
  \vspace{-10pt}
\end{table*}

\section{Experimental Setup}

\subsection{Protocols}
\label{ssec:protocols}

\UXBench evaluates judge models using two complementary protocols: an automated LLM-as-judge sweep and a blind human validation study.

\noindent \textbf{Automated sweep.}
Each evaluated model drives the same coverage-gated exploration agent over the runnable fixtures in \autoref{tab:catalog} under a fixed exploration budget, producing one complete evaluation run per model--fixture pair. Each run is converted into an evidence-grounded UX report and scored by a fixed scoring judge, GPT-5.4-Mini. The fixed scoring judge is blinded to the identity of the judge model that produced each repair report.

\begin{wrapfigure}{r}{0.5\textwidth}
  \centering
  \includegraphics[width=\linewidth]{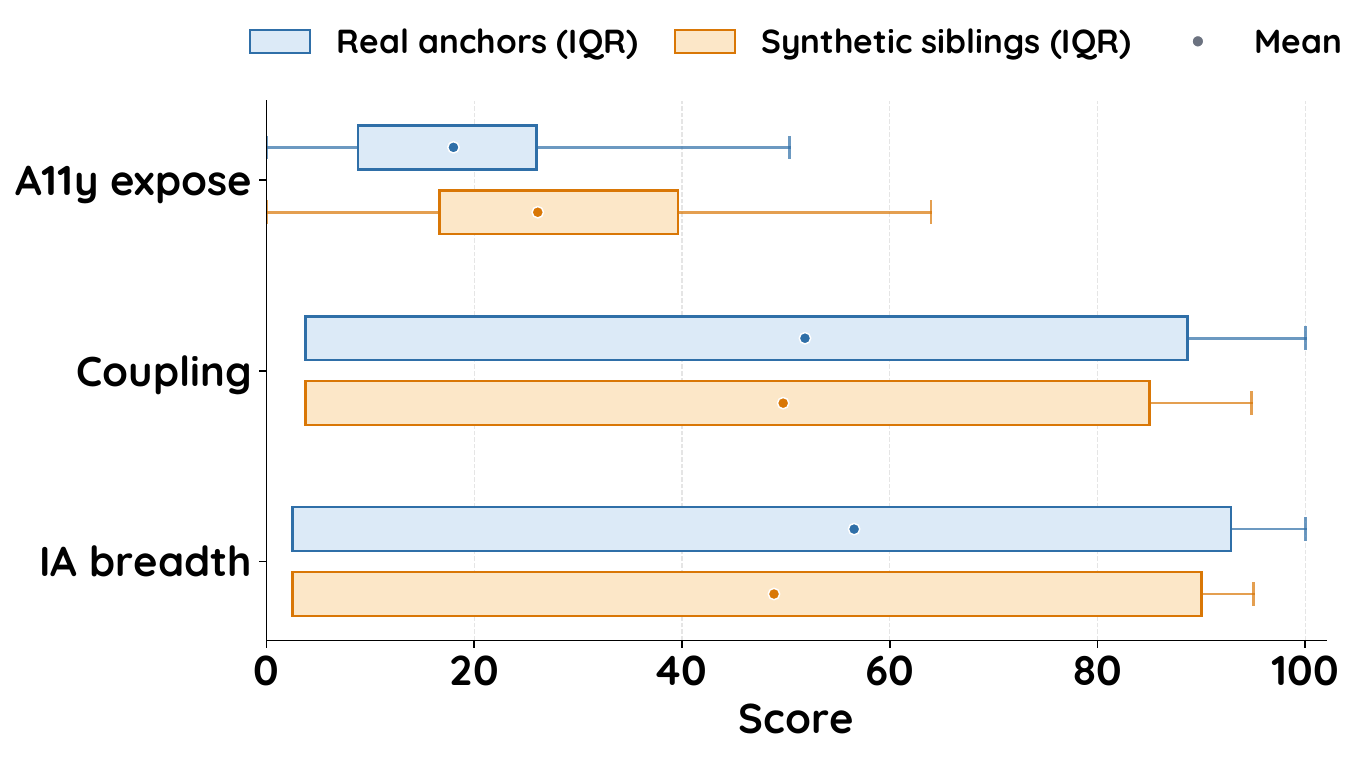}
  \caption{
  Structural characterization of real anchors and synthetic siblings in \UXBench. A11y denotes accessibility, and IA denotes information architecture.
  }
  \label{fig:dataset_characterization}
\end{wrapfigure}

\noindent \textbf{Human evaluation.}
To validate whether automated repair-lift rankings align with human UX judgment, we conduct a blind expert review of the repaired candidate interfaces. We recruit six participants with UX or front-end design experience. For each fixture, reviewers inspect the rendered repaired webpage produced from an anonymized judge report and rate the interface on the same seven rubric dimensions, with model identities, presentation order, and automated scores hidden throughout. Reviewers may revisit reports within the same group before submission, allowing local recalibration across candidates.

\subsection{Scoring Rubric}
\label{ssec:rubric}

\UXBench scores each fixture along the seven default dimensions in \autoref{tab:rubric}. The rubric reuses constructs measured by the standard family of validated UX questionnaires (e.g., SUS, UEQ, and NASA-TLX; details in \autoref{app:rubricConstruction}), but re-grounds each construct as a criterion that can be answered from the interaction trace rather than from a post-task self-report. Each metric is rated on a 1--5 ordinal scale (5 indicates no meaningful issue, 1 a blocking failure), and the rubric score is the equally weighted mean of its metrics.

\subsection{Dataset Characterization.}
\UXBench contains 41fixtures: 11 real anchors and 30 synthetic siblings. \autoref{fig:dataset_characterization} characterizes \UXBench fixtures along three structural axes. The clearest separation appears in accessibility exposure: synthetic siblings have a higher interquartile range and mean than real anchors, suggesting that they introduce more localized implementation-level risks such as missing labels, weak accessibility cues, or placeholder interactions. In contrast, real anchors show higher information-architecture breadth, with both the mean and upper range shifted slightly upward, reflecting broader navigation structures and more system-like page organization. Cross-page coupling does not clearly separate real and synthetic fixtures: both groups span almost the full score range and have highly overlapping distributions. This overlap suggests that coupling is not primarily determined by whether a fixture is real or synthetic, but instead reflects within-benchmark heterogeneity. Overall, the figure shows that \UXBench is not organized along a single difficulty axis; it combines fixtures that stress global reasoning over navigation and state consistency with fixtures that stress local inspection of feedback, accessibility, and recovery behavior.

\begin{table*}[t]
  \centering
  \small
  \setlength{\tabcolsep}{2.3pt}
  \renewcommand{\arraystretch}{1.28}
  \begin{tabularx}{\textwidth}{
    @{}
    >{\raggedright\arraybackslash}p{0.16\textwidth}
    >{\centering\arraybackslash}X
    >{\centering\arraybackslash}X
    >{\centering\arraybackslash}X
    >{\centering\arraybackslash}X
    >{\centering\arraybackslash}X
    >{\centering\arraybackslash}X
    >{\centering\arraybackslash}X
    >{\centering\arraybackslash}X
    >{\centering\arraybackslash}X
    >{\centering\arraybackslash}X
    @{}
  }
    \toprule
    \textbf{Model}
    & \textbf{Land.}
    & \textbf{Price}
    & \textbf{Onbd.}
    & \textbf{Book.}
    & \textbf{Dash}
    & \textbf{Docs}
    & \textbf{Privacy}
    & \textbf{Visual.}
    & \textbf{Chatbot}
    & \textbf{Mobile} \\
    \midrule

    \modelcell{GPT 5.4}{openai}
    & \textbf{0.667} & 0.714 & 0.375 & 0.411 & 0.583 & \textbf{0.738} & 0.589 & 0.333 & 0.417 & 0.446 \\

    \rowcolor{RowGray}
    \modelcell{Kimi k2.5}{kimi}
    & 0.637 & 0.518 & 0.429 & 0.661 & 0.381 & 0.637 & 0.464 & \textbf{0.786} & 0.643 & 0.286 \\

    \modelcell{Sonnet 4.6}{claude}
    & 0.476 & 0.196 & 0.536 & 0.393 & 0.589 & 0.690 & 0.250 & 0.381 & \textbf{0.851} & 0.411 \\

    \rowcolor{RowGray}
    \modelcell{GPT Mini}{openai}
    & 0.452 & 0.661 & 0.521 & 0.536 & 0.292 & 0.286 & \textbf{0.643} & 0.571 & 0.500 & 0.536 \\

    \modelcell{GLM 5.1}{glm}
    & 0.411 & 0.393 & 0.543 & 0.536 & 0.625 & 0.560 & 0.554 & 0.595 & 0.417 & \textbf{0.714} \\

    \rowcolor{RowGray}
    \modelcell{GPT Nano}{openai}
    & 0.429 & \textbf{0.804} & \textbf{0.664} & \textbf{0.768} & 0.476 & 0.250 & 0.518 & 0.262 & 0.482 & 0.536 \\

    \modelcell{Qwen 3.6}{qwen}
    & 0.548 & 0.500 & 0.461 & 0.482 & 0.363 & 0.315 & 0.571 & 0.524 & 0.381 & 0.571 \\

    \rowcolor{RowGray}
    \modelcell{Gemini 3.1}{gemini}
    & 0.381 & 0.214 & 0.471 & 0.214 & \textbf{0.690} & 0.524 & 0.411 & 0.548 & 0.310 & 0.500 \\

    \bottomrule
  \end{tabularx}
  \caption{\label{tab:category_winrate}
  Category-level pairwise win rates across the ten \UXBench surface families. 
  Land., Price, Onbd., Book., and Dash denote Landing, Pricing, Onboarding, Booking, and Dashboard, respectively; 
  Visual. denotes visual design. Each cell reports a model's mean pairwise win rate against the other seven evaluated models within the same surface family. 
  \textbf{Bold} indicates the highest win rate in each category.}
    \vspace{-10pt}
\end{table*}

\begin{figure*}[t]
  \centering
  \includegraphics[width=\textwidth]{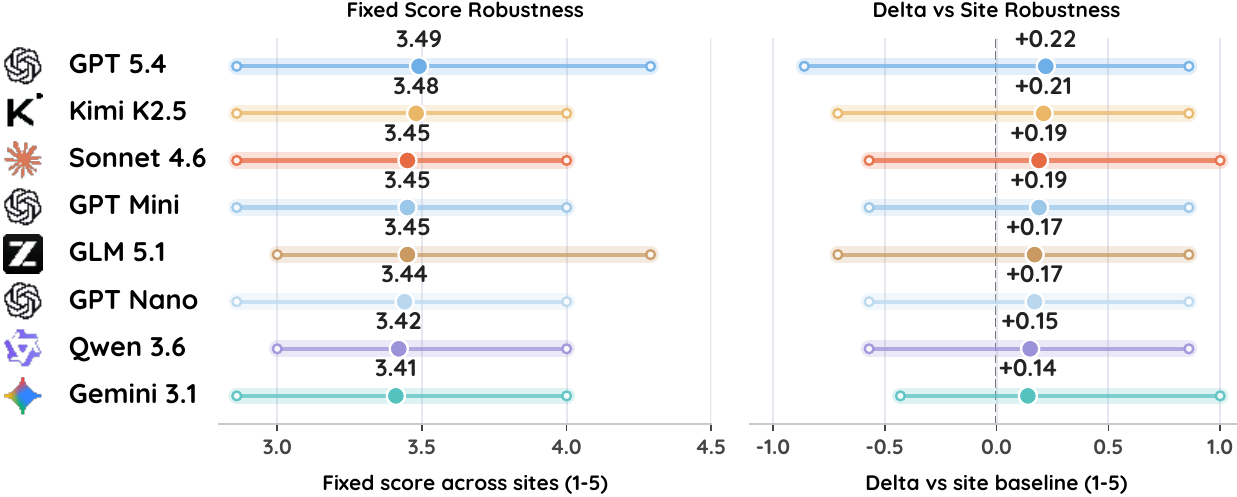}
  \caption{
  Robustness across judge models. Per-fixture repaired scores and uplift $\Delta$ relative to the site baseline. Models with higher mean lift exhibit wider site-level spreads, separating the sweep along an axis the aggregate column cannot expose.
  }
  \label{fig:robustness}
  \vspace{-10pt}
\end{figure*}

\begin{figure*}[t]
  \centering
  \includegraphics[width=\textwidth]{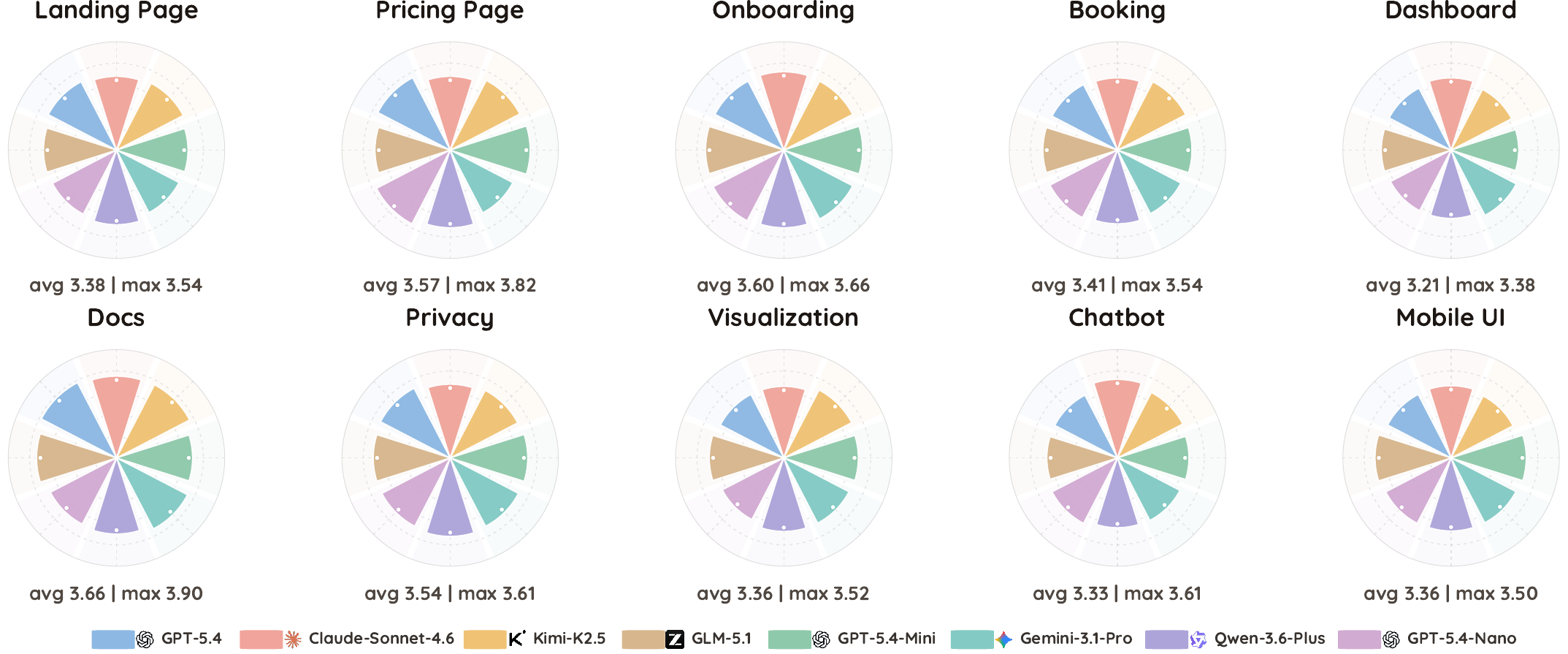}
  \caption{\label{fig:category}
  Category-level repaired-score profiles across the eight evaluated models. Each panel reports the category mean and the best repaired score. The leading model changes across surface types, and the gap between category averages and best-model scores varies noticeably across surfaces.
  }
  \vspace{-10pt}
\end{figure*}

\section{Results}
\label{sec:results}

\noindent \omark{1} \textbf{Judge models differ in their ability to produce repair-actionable UX reports.}
\autoref{tab:overall} shows that \UXBench exposes measurable differences among judge models under a paired evaluation protocol. Since every model is evaluated on the same fixture set, repaired by the same fixed agent, and scored by the same evaluator, the remaining variation reflects differences in the reports produced by the judge models. Although all models improve over the unrepaired baseline, the size of the improvement differs across judges: GPT-5.4 obtains the largest repair lift (+0.22), while Gemini-3.1-Pro obtains the smallest (+0.14), yielding an 0.08-point spread on the 1--5 rubric scale. This gap indicates that UX report generation is not saturated among current frontier models: models are not interchangeable as UX judges, even when downstream repair and scoring are held fixed. At the same time, the narrow absolute spread suggests that adjacent ranks should be interpreted cautiously and analyzed together with fixture-level variation rather than treated as a definitive global ordering. \autoref{app:stat-validation} reports site-paired significance tests, bootstrap confidence intervals, and effect sizes for repair-lift estimates.

\begin{wrapfigure}{r}{0.48\columnwidth}
  \centering
  \includegraphics[width=\linewidth]{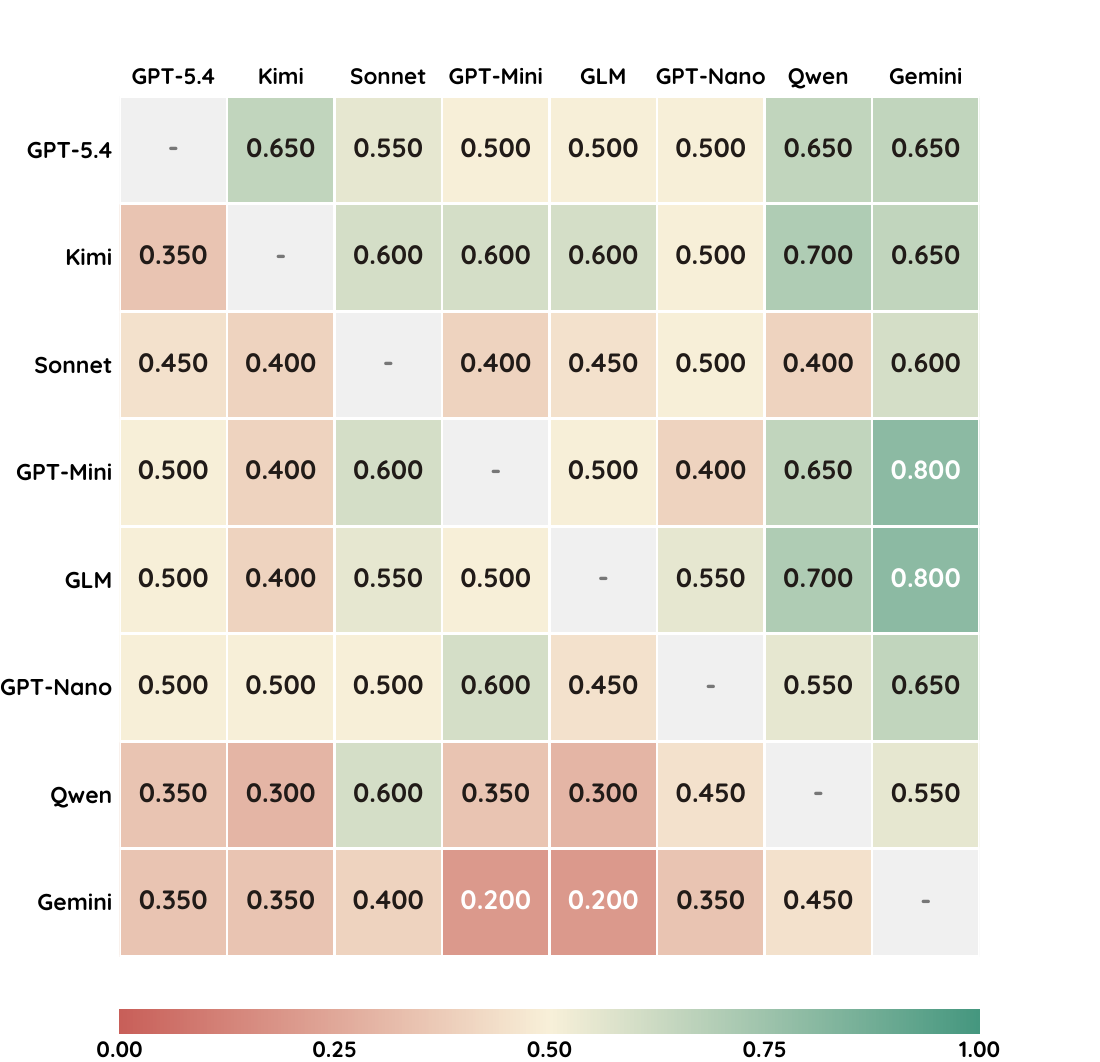}
  \caption{
  Pairwise model wins. Overall pairwise win-rate matrix across evaluated models.
  }
  \label{fig:winrate_heatmap}
\end{wrapfigure}

\noindent \omark{2} \textbf{Judge models exhibit distinct rubric-level repair signatures.}
The per-dimension deltas in \autoref{tab:overall} show that models differ not only in the amount of repair lift they produce, but also in the UX dimensions where that lift appears. GPT-5.4 obtains the strongest gain on error recovery, Kimi-K2.5 leads on feedback and trust transparency, Claude-Sonnet-4.6 is strongest on goal-state clarity, and Qwen-3.6-Plus reaches the largest gains on flow and scanability/accessibility. These model-specific signatures indicate that UX report quality is not a one-dimensional ability: two models with similar aggregate scores may still support different kinds of downstream repair.

\noindent \omark{3} \textbf{Judge models vary in fixture-level reliability.}
\autoref{fig:robustness} shows that models also differ in how consistently their reports transfer across individual fixtures. GPT-5.4 achieves the highest mean repaired score and mean lift, but its site-level interval is wide, indicating uneven performance across the benchmark. Some lower-mean models show tighter site-level distributions, suggesting more stable but less peak-performing behavior. This variation matters in practice: a judge can produce highly actionable reports on some interfaces while yielding weak repair outcomes on others.

\noindent \omark{4} \textbf{UX repair decomposes into surface-conditioned subcompetences.}
\autoref{fig:category} shows that surface families differ in both the leading model and the headroom they expose. Documentation and pricing pages reach relatively high repaired scores, whereas dashboard and chatbot/agent interfaces remain lower on average, suggesting that some surfaces are easier to improve once actionable issues are identified, while others require harder diagnosis of dense state, interaction history, or conversational feedback. \autoref{tab:category_winrate} provides a complementary pairwise view: the strongest model in one category often fails to dominate in another, and winning margins vary across columns rather than following the aggregate ranking. Thus, category-level variation is not merely a visualization artifact of mean scores; it reflects surface-conditioned differences in what UX evidence a judge can collect, prioritize, and translate into actionable findings. \UXBench should therefore be read as a multi-granularity benchmark that separates global report quality from surface-specific diagnostic competence.

\noindent \omark{5} \textbf{Pairwise win rates expose the stability of model separation.}
Mean repaired scores summarize average report actionability, but they do not show whether a model wins broadly across fixtures or is driven by a few large gains. \autoref{fig:winrate_heatmap} addresses this by converting overall performance into paired fixture-level comparisons. The matrix reveals a top cluster rather than a single dominant judge: GPT-5.4, Kimi-K2.5, and GLM-5.1 each average about 0.57 pairwise win rate against the other models, while several head-to-head cells among nearby models remain at or near 0.5. This means that aggregate leaders are separated, but not by a strict total order. Conversely, Gemini-3.1-Pro shows a broadly weak pairwise profile, losing to most alternatives even though it can still lead a specific surface family such as dashboards. Pairwise win rates therefore refine the aggregate results: \UXBench does not merely rank models by mean repair lift, but reveals whether those advantages persist across heterogeneous interfaces or depend on localized strengths.

\begin{figure*}[t]
  \centering
  \includegraphics[width=\textwidth]{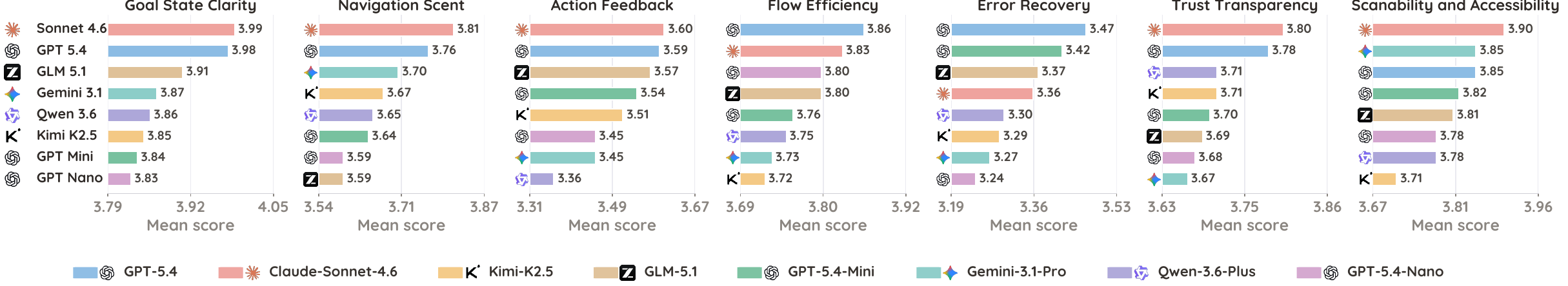}
  \caption{\label{fig:human}
    Human-rated UX profiles of repaired interfaces produced from each judge model's report. Each panel shows the mean human score for one rubric dimension under blind model identity.
  }
  \vspace{-10pt}
\end{figure*}

\noindent \textbf{\omark{6} Human evaluation confirms the broad signal but sharpens the top-tier separation.}
\autoref{fig:human} reports blind human ratings of the repaired interfaces produced from different judge-model reports. The human results broadly support the automated protocol at the coarse level: interfaces repaired from stronger judge reports are generally perceived as more usable by expert reviewers. However, the human ratings also make the separation less like a strict leaderboard and more like a top-tier cluster. GPT-5.4 and Claude-Sonnet-4.6 form the clearest upper group, with small dimension-level differences rather than a uniformly dominant model across all criteria. Meanwhile, some models that are competitive under automated repair lift do not translate that advantage into equally strong perceived interface quality. This distinction is important because automated repair lift measures whether a report helps a fixed repair agent improve rubric scores, whereas human evaluation captures whether the resulting webpage feels clearer, easier to navigate, more responsive, and more trustworthy to reviewers. Thus, the human study calibrates \UXBench by supporting its actionability signal while showing that close model comparisons should be interpreted through human-perceived interface quality rather than automated scores alone. More detailed human--LLM comparison results are provided in \autoref{app:eval_results}.
\section{Conclusion}
We proposed \UXBench, a benchmark for evaluating LLMs as interaction-grounded UX judges. \UXBench combines runnable fixtures, coverage-gated exploration, evidence-grounded reports, and report-conditioned repair to test whether model critiques can drive real interface improvements. Our results show that UX judging remains unsaturated and multi-dimensional, with models differing across actionability, reliability, rubric dimensions, and product surfaces.

\clearpage

\section*{Limitations}
This work has several limitations. First, \UXBench uses local-first static fixtures to ensure control and reproducibility, but this design cannot fully capture the dynamics of live production systems, such as personalization, backend failures, or long-term user behavior. Second, report actionability is measured through a fixed repair agent and fixed scorer, so the results reflect how well reports support this controlled repair pipeline rather than all possible developer workflows. Finally, our human validation uses expert reviewers but remains limited in scale; broader studies with more diverse users and deployment contexts would further clarify how automated UX-judge signals translate to real-world UX improvement.

\section*{Ethical Considerations}

This work evaluates LLMs as UX judges in controlled, local-first web fixtures rather than on live user-facing services. The benchmark does not require user accounts, collect personal user data, or interact with production systems. Human validation is conducted through blind review of anonymized model reports, with model identities and automated scores hidden from reviewers. A potential risk is that automated UX judges may be over-trusted or used as substitutes for human-centered design practice. We emphasize that \UXBench measures report actionability under a controlled repair pipeline and should complement, rather than replace, human usability studies, accessibility audits, and domain-specific expert review.

\clearpage
\bibliographystyle{plainnat}
\bibliography{references}

\clearpage
\appendix
\onecolumn

\section{Details of Coverage Gate Algorithm}
\label{app:algoDetail}

\begin{algorithm}[ht]
\caption{Coverage-Gated Exploration}
\label{alg:algo_detailed}
\footnotesize
\begin{algorithmic}[1]
\Require fixture profile $P$ with target pages $\mathcal{U}$ and prescanned controls $\mathcal{K}$
\Require exploration plan $\Pi$, step budget $B$
\Require depth $d\in\{\textsc{quick},\textsc{std},\textsc{deep},\textsc{exh}\}$
\Require viewport mode $v\in\{\textsc{desk},\textsc{mob},\textsc{both}\}$
\Ensure exploration trace $\tau$ and evidence confidence $q\in\{\textsc{low},\textsc{med},\textsc{high}\}$

\State $\mathcal{V}\gets\textsc{Viewports}(v)$
\State $\mathcal{G}\gets\textsc{CoverageGoals}(P,\Pi,d,\mathcal{V})$
\State open the first target page in $\Pi$ under the first viewport in $\mathcal{V}$
\State $o_0\gets\textsc{Observe}()$; $\tau\gets[o_0]$

\For{$t=0,\ldots,B-1$}
    \State compute $\mathcal{P}_{\mathrm{seen}}$, $\mathcal{V}_{\mathrm{seen}}$,
    $\mathcal{K}_{\mathrm{done}}$, and $\mathcal{S}_{\mathrm{seen}}$ from $\tau$

    \State $C.\textit{unmet}\gets
        (\mathcal{G}_{\mathrm{page}}\setminus\mathcal{P}_{\mathrm{seen}})
        \cup
        (\mathcal{G}_{\mathrm{viewport}}\setminus\mathcal{V}_{\mathrm{seen}})$
    \Statex \hspace{\algorithmicindent}$\phantom{C.\textit{unmet}\gets{}}
        \cup
        (\mathcal{G}_{\mathrm{control}}\setminus\mathcal{K}_{\mathrm{done}})
        \cup
        (\mathcal{G}_{\mathrm{state}}\setminus\mathcal{S}_{\mathrm{seen}})$

    \State $C.\textit{stop\_allowed}\gets
        (C.\textit{unmet}=\emptyset
        \wedge
        \textsc{NoBlockingFailure}(\tau))$

    \State $a_t\gets\textsc{Brain}(o_t,\tau,\Pi,C)$

    \If{$a_t=\textsc{finish}$ \textbf{and} $C.\textit{stop\_allowed}=\bot$}
        \State $a_t\gets\textsc{Brain}(o_t,\tau,\Pi,C,$
        \Statex \hspace{\algorithmicindent}$\phantom{a_t\gets\textsc{Brain}(}
            \textit{must\_continue}=C.\textit{unmet})$
    \EndIf

    \If{$a_t=\textsc{finish}$ \textbf{and} $C.\textit{stop\_allowed}=\bot$}
        \State $a_t\gets\textsc{Fallback}(C)$
    \EndIf

    \If{$a_t=\textsc{finish}$}
        \State \textbf{break}
    \EndIf

    \If{$\textsc{Unsafe}(a_t,P)$}
        \State $a_t\gets\textsc{Fallback}(C)$
    \EndIf

    \State execute $a_t$
    \State $o_{t+1}\gets\textsc{Observe}()$
    \State append $(a_t,o_{t+1})$ to $\tau$
\EndFor

\State recompute all coverage sets and $C$ from $\tau$

\If{$C.\textit{stop\_allowed}=\top$}
    \State $q\gets\textsc{high}$
\ElsIf{$\mathcal{G}_{\mathrm{page}}\subseteq\mathcal{P}_{\mathrm{seen}}$
    \textbf{and}
    $\mathcal{G}_{\mathrm{viewport}}\subseteq\mathcal{V}_{\mathrm{seen}}$}
    \State $q\gets\textsc{med}$
\Else
    \State $q\gets\textsc{low}$
\EndIf

\State \Return $\tau,q$
\end{algorithmic}
\end{algorithm}

\begin{table*}[ht]
  \centering
  \small
  \setlength{\tabcolsep}{4.5pt}
  \renewcommand{\arraystretch}{1.16}

  \rowcolors{2}{RowGray}{white}
  \begin{tabularx}{\textwidth}{
    @{}
    >{\centering\arraybackslash}p{0.04\textwidth}
    >{\raggedright\arraybackslash}p{0.15\textwidth}
    >{\columncolor{AnchorBlue}\raggedright\arraybackslash}p{0.27\textwidth}
    >{\columncolor{SiblingGreen}\raggedright\arraybackslash}X
    @{}
  }
    \toprule
    \textbf{\#} & \textbf{Category} & \textbf{Real anchor} & \textbf{Synthetic siblings} \\
    \midrule

    1  & Landing Page
       & \site{notion}
       & \site{pelagic}, \site{meadowos}, \site{stratabox} \\

    2  & Pricing Page
       & \site{slack}
       & \site{codekite}, \site{lattice}, \site{vaultkey} \\

    3  & Onboarding
       & \site{shopify}, \site{govuk-passport}
       & \site{solstice-bank}, \site{greengrove}, \site{civicport} \\

    4  & Booking
       & \site{booking}
       & \site{orbitride}, \site{moonlight-tickets}, \site{tablerose} \\

    5  & Dashboard
       & \site{cloudflare-radar}
       & \site{fleetatlas}, \site{aeroiq}, \site{pulsegrid} \\

    6  & Docs
       & \site{stripe-docs}
       & \site{tessera}, \site{weaveapi}, \site{runeforge-docs} \\

    7  & Privacy
       & \site{microsoft-privacy}
       & \site{privacy-dashboard}, \site{aurora-network}, \site{meadowid} \\

    8  & Visualization
       & \site{owid-population}
       & \site{fred-unrate}, \site{climate-almanac}, \site{migration-atlas} \\

    9  & ChatBot
       & \site{chatgpt}
       & \site{lumen-research}, \site{forge-coder}, \site{atlas-tutor} \\

    10 & Mobile UI
       & \site{ridenow}
       & \site{brewlog}, \site{harborwallet}, \site{larkfit} \\

    \bottomrule
  \end{tabularx}

  \caption{\label{tab:catalog}
  \UXBench site catalog. Real anchors ground each category in recognizable public-product interaction patterns, while synthetic siblings preserve the core interaction model with independently authored branding, content, layout, and visual identity.}
\end{table*}



\section{Surface Categories}
\label{app:surfaceCategory}

\paragraph{Fixture Design and Catalog.}
\UXBench is built from local-first web fixtures implemented as static HTML/CSS/JavaScript bundles. Each fixture can be served from a local file server and evaluated without accounts, backend services, third-party APIs, or live-site dependencies. This design makes the benchmark reproducible: model comparisons are not affected by A/B tests, personalization, network failures, or changes in production websites. At the same time, the fixtures remain interactive rather than screenshot-only, allowing judge models to click controls, enter text, switch viewports, observe feedback, and ground their reports in actual interface behavior.

As shown in \autoref{tab:catalog}, \UXBench spans ten product-surface families, covering common web experiences such as landing pages, pricing pages, onboarding flows, booking interfaces, dashboards, documentation, privacy/settings pages, data visualizations, chatbot interfaces, and mobile micro-UIs. Within each family, we pair real-product anchors with independently authored synthetic siblings. The real anchors ground each category in recognizable public-product interaction patterns, while the synthetic siblings preserve the same broad interaction model but vary branding, copy, layout, and visual identity. This anchor--sibling structure balances external validity with control: anchors expose models to realistic UX conventions, while siblings test whether a judge evaluates the current interface rather than relying on memorized impressions of familiar products.

\section{Rubric Construction}
\label{app:rubricConstruction}

\UXBench derives its rubric from established usability and user-experience instruments, but does not directly administer these instruments as questionnaires. Instead, as summarized in \autoref{tab:rubric}, we use them as construct references and translate recurring notions such as clarity, navigation support, feedback, effort, recovery, trust, and scanability into browser-grounded evaluation questions. This lets the rubric preserve links to prior UX measurement practice while remaining answerable from interaction evidence collected in a live web interface.

\paragraph{System Usability Scale (SUS).}
SUS provides a compact reference for perceived usability, learnability, consistency, and user confidence~\citep{brooke1996sus}. In \UXBench, these constructs mainly support goal-state clarity and trust/consequence transparency.

\paragraph{User Experience Questionnaire (UEQ).}
UEQ covers both pragmatic and hedonic aspects of user experience, including perspicuity, efficiency, dependability, stimulation, and attractiveness~\citep{laugwitz2008ueq}. We primarily use its pragmatic dimensions to motivate clarity, flow efficiency, and scanability.

\paragraph{Post-Study System Usability Questionnaire (PSSUQ).}
PSSUQ emphasizes system usefulness, information quality, and interface quality~\citep{lewis1995ibm,lewis2002pssuq}. These constructs inform our treatment of navigation cues, action feedback, recovery guidance, and task-supporting information.

\paragraph{Standardized User Experience Percentile Rank Questionnaire (SUPR-Q).}
SUPR-Q is especially relevant to website-level evaluation because it covers usability, trust, appearance, and loyalty~\citep{sauro2015suprq}. We use it to ground navigation scent, trust transparency, and scanability for web interfaces.

\paragraph{UMUX-Lite.}
UMUX-Lite reduces perceived usability to whether the system meets user requirements and is easy to use~\citep{lewis2013umuxlite}. These two judgments align with goal fit, discoverability, and confidence in continuing.

\begin{table}[t]
  \centering
  \small
  \setlength{\tabcolsep}{8pt}
  \renewcommand{\arraystretch}{1.18}

  \begin{tabularx}{0.82\linewidth}{
    @{}
    >{\raggedright\arraybackslash}p{0.42\linewidth}
    >{\raggedright\arraybackslash}X
    >{\centering\arraybackslash}p{0.12\linewidth}
    @{}
  }
    \toprule
    \rowcolor{HeaderGray}
    \textbf{Model} & \textbf{Provider} & \textbf{Open} \\
    \midrule

    \modelcell{GPT-5.4}{openai}
      & OpenAI & \ding{55} \\

    \rowcolor{RowGray}
    \modelcell{Kimi-K2.5}{kimi}
      & Moonshot AI & \ding{51} \\

    \modelcell{Claude-Sonnet-4.6}{claude}
      & Anthropic & \ding{55} \\

    \rowcolor{RowGray}
    \modelcell{GPT-5.4-Mini}{openai}
      & OpenAI & \ding{55} \\

    \modelcell{GLM-5.1}{glm}
      & Zhipu AI & \ding{51} \\

    \rowcolor{RowGray}
    \modelcell{GPT-5.4-Nano}{openai}
      & OpenAI & \ding{55} \\

    \modelcell{Qwen-3.6-Plus}{qwen}
      & Alibaba & \ding{55} \\

    \rowcolor{RowGray}
    \modelcell{Gemini-3.1-Pro}{gemini}
      & Google & \ding{55} \\

    \bottomrule
  \end{tabularx}

  \caption{\label{tab:evaluated_models}
  Judge models used in the \UXBench benchmark sweep. \ding{51} indicates open-weight availability; \ding{55} indicates API-only access.}
  \vspace{-10pt}
\end{table}

\paragraph{Single Ease Question (SEQ).}
SEQ captures the perceived difficulty of a completed task~\citep{sauro2009seq}. We use it as a task-local reference for flow efficiency, especially when a path requires unnecessary steps, repeated input, or backtracking.

\paragraph{After-Scenario Questionnaire (ASQ).}
ASQ focuses on satisfaction with task completion, time required, and supporting information~\citep{lewis1991asq}. It motivates our use of action feedback, flow efficiency, and error recovery as interaction-path properties.

\paragraph{Customer Effort Score (CES).}
CES highlights the effort users must expend to accomplish a goal or resolve a problem~\citep{dixon2010stop}. In \UXBench, it supports our attention to friction, repeated work, hidden requirements, and avoidable detours.

\paragraph{NASA Task Load Index (NASA-TLX).}
NASA-TLX provides vocabulary for workload, effort, time pressure, and frustration~\citep{hart1988nasatlx}. We use it to capture cognitive burden in dense flows, unclear states, and difficult recovery paths.

\clearpage

\begin{table*}[t]
  \centering
  \setlength{\tabcolsep}{4pt}
  \renewcommand{\arraystretch}{1.22}

  \begin{tabularx}{\textwidth}{
    @{}
    >{\centering\arraybackslash}p{0.025\textwidth}
    >{\raggedright\arraybackslash}p{0.19\textwidth}
    >{\raggedright\arraybackslash}X
    >{\raggedright\arraybackslash}p{0.10\textwidth}
    >{\raggedright\arraybackslash}p{0.27\textwidth}
    @{}
  }
    \toprule
    \textbf{\#} & \textbf{\UXBench metric} & \textbf{In-browser evaluation question} & \textbf{Reference scales} & \textbf{Operationalized constructs} \\
    \midrule

    1 & Goal-state clarity
      & Can users quickly understand the page purpose, current state, available options, and most sensible next action?
      & UEQ, PSSUQ, SUS, UMUX-Lite
      & Perspicuity; information quality; perceived ease of use; confidence in using the system \\

    \rowcolor{RowGray}
    2 & Navigation scent
      & Do labels, menus, tabs, search, and filters provide reliable cues toward the right content or next step?
      & SUPR-Q, PSSUQ, UEQ
      & Ease of navigation; ease of finding information; clarity of organization; learnability cues \\

    3 & Action feedback
      & Are user actions followed by clear feedback for selection, input, loading, validation, success, and failure states?
      & ASQ, PSSUQ, NASA-TLX
      & Ease of task completion; support information; system-status visibility; error feedback; frustration \\

    \rowcolor{RowGray}
    4 & Flow efficiency
      & Can users complete multi-step or cross-page tasks without unnecessary detours, repetition, waiting, or backtracking?
      & SEQ, ASQ, CES, UEQ
      & Task ease; time satisfaction; perceived effort; efficiency; cognitive burden \\

    5 & Error recovery
      & Does the interface prevent likely mistakes and provide clear ways to correct, undo, retry, or return when problems occur?
      & PSSUQ, ASQ, NASA-TLX
      & Error prevention; recovery from mistakes; helpful guidance; support information; effort and frustration \\

    \rowcolor{RowGray}
    6 & Trust transparency
      & Before committing, can users understand costs, permissions, privacy-relevant choices, and consequences of sensitive actions?
      & SUPR-Q, SUS, PSSUQ, UMUX-Lite
      & Trust; confidence transacting; confidence using the system; clarity of consequences; information quality \\

    7 & Scanability and accessibility
      & Is the page easy to scan, visually prioritized, readable across screen sizes, and operable with basic accessibility cues?
      & SUPR-Q, UEQ, PSSUQ
      & Appearance; attractiveness; readability; organized information; responsive usability; basic accessibility support \\

    \bottomrule
  \end{tabularx}

  \caption{\label{tab:rubric}
  \UXBench default scoring rubric and construct alignment. The seven metrics operationalize observable interface qualities that recur across established UX instruments, including SUS, UEQ, PSSUQ, SUPR-Q, UMUX-Lite, SEQ, ASQ, CES, and NASA-TLX, but restate them as browser-grounded questions answerable from interaction evidence rather than post-task self reports.}
\end{table*}

\clearpage

\section{Evaluated Judge Models}
\label{app:evaluated_models}

\UXBench evaluates eight frontier LLMs as UX judge models: GPT-5.4, GPT-5.4-Mini, GPT-5.4-Nano~\citep{openai2026gpt54,openai2026gpt54mininano}, Claude-Sonnet-4.6~\citep{anthropic2026claudesonnet46}, Gemini-3.1-Pro~\citep{google2026gemini31pro}, Kimi-K2.5~\citep{moonshot2026kimi25}, GLM-5.1~\citep{zai2026glm51}, and Qwen-3.6-Plus~\citep{qwen2026qwen36plus}. These models span major closed API systems and competitive open or open-weight-oriented model families, covering recent progress in agentic reasoning, coding, multimodal understanding, and computer-use capabilities. In our benchmark, each model is used only as a judge/report generator: it inspects the same web fixtures, produces UX reports under the same rubric, and is compared through the same downstream repair-and-scoring protocol. \autoref{tab:evaluated_models} summarizes the evaluated models and their providers.

\section{LLM-Human Judge Comparison}
\label{app:eval_results}

\begin{table*}[t]
  \centering
  \small
  \setlength{\tabcolsep}{2.3pt}
  \renewcommand{\arraystretch}{1.28}
  \begin{tabularx}{\textwidth}{
    @{}
    >{\raggedright\arraybackslash}p{0.16\textwidth}
    >{\centering\arraybackslash}X
    >{\centering\arraybackslash}X
    >{\centering\arraybackslash}X
    >{\centering\arraybackslash}X
    >{\centering\arraybackslash}X
    >{\centering\arraybackslash}X
    >{\centering\arraybackslash}X
    >{\centering\arraybackslash}X
    >{\centering\arraybackslash}X
    >{\centering\arraybackslash}X
    @{}
  }
    \toprule
    \textbf{Model}
    & \textbf{Land.}
    & \textbf{Price}
    & \textbf{Onbd.}
    & \textbf{Book.}
    & \textbf{Dash}
    & \textbf{Docs}
    & \textbf{Privacy}
    & \textbf{Visual.}
    & \textbf{Chatbot}
    & \textbf{Mobile} \\
    \midrule

    \modelcell{GPT 5.4}{openai}
    & \textbf{0.696} & \textbf{0.804} & \textbf{0.886} & 0.732 & 0.696 & 0.696 & 0.482 & 0.786 & 0.554 & \textbf{0.821} \\

    \rowcolor{RowGray}
    \modelcell{Kimi k2.5}{kimi}
    & 0.393 & 0.304 & 0.614 & 0.339 & 0.429 & 0.196 & 0.464 & 0.452 & 0.321 & 0.321 \\

    \modelcell{Sonnet 4.6}{claude}
    & 0.607 & 0.607 & 0.443 & 0.732 & \textbf{0.857} & 0.643 & \textbf{0.768} & \textbf{0.905} & 0.571 & 0.714 \\

    \rowcolor{RowGray}
    \modelcell{GPT Mini}{openai}
    & 0.286 & 0.429 & 0.371 & 0.446 & 0.429 & \textbf{0.732} & 0.339 & 0.429 & 0.714 & 0.589 \\

    \modelcell{GLM 5.1}{glm}
    & 0.446 & 0.268 & 0.400 & 0.339 & 0.446 & 0.482 & 0.554 & 0.524 & \textbf{0.857} & 0.554 \\

    \rowcolor{RowGray}
    \modelcell{GPT Nano}{openai}
    & 0.500 & 0.643 & 0.543 & 0.304 & 0.232 & 0.286 & 0.375 & 0.333 & 0.375 & 0.339 \\

    \modelcell{Qwen 3.6}{qwen}
    & 0.518 & 0.464 & 0.257 & 0.357 & 0.321 & 0.393 & 0.375 & 0.262 & 0.196 & 0.286 \\

    \rowcolor{RowGray}
    \modelcell{Gemini 3.1}{gemini}
    & 0.554 & 0.482 & 0.486 & \textbf{0.750} & 0.589 & 0.571 & 0.643 & 0.310 & 0.411 & 0.375 \\

    \bottomrule
  \end{tabularx}
  \caption{\label{tab:human_category_winrate}
  Category-level human pairwise win rates across the ten \UXBench surface families.
  Land., Price, Onbd., Book., and Dash denote Landing, Pricing, Onboarding, Booking, and Dashboard, respectively;
  Visual. denotes visual design. Each cell reports a model's mean pairwise win rate against the other evaluated models within the same surface family.
  \textbf{Bold} indicates the highest human-preferred model in each category.}
  \vspace{-10pt}
\end{table*}

\section{Human Evaluation Details}
\label{app:human_eval}

\begin{figure*}[t]
  \centering
  \includegraphics[width=\linewidth]{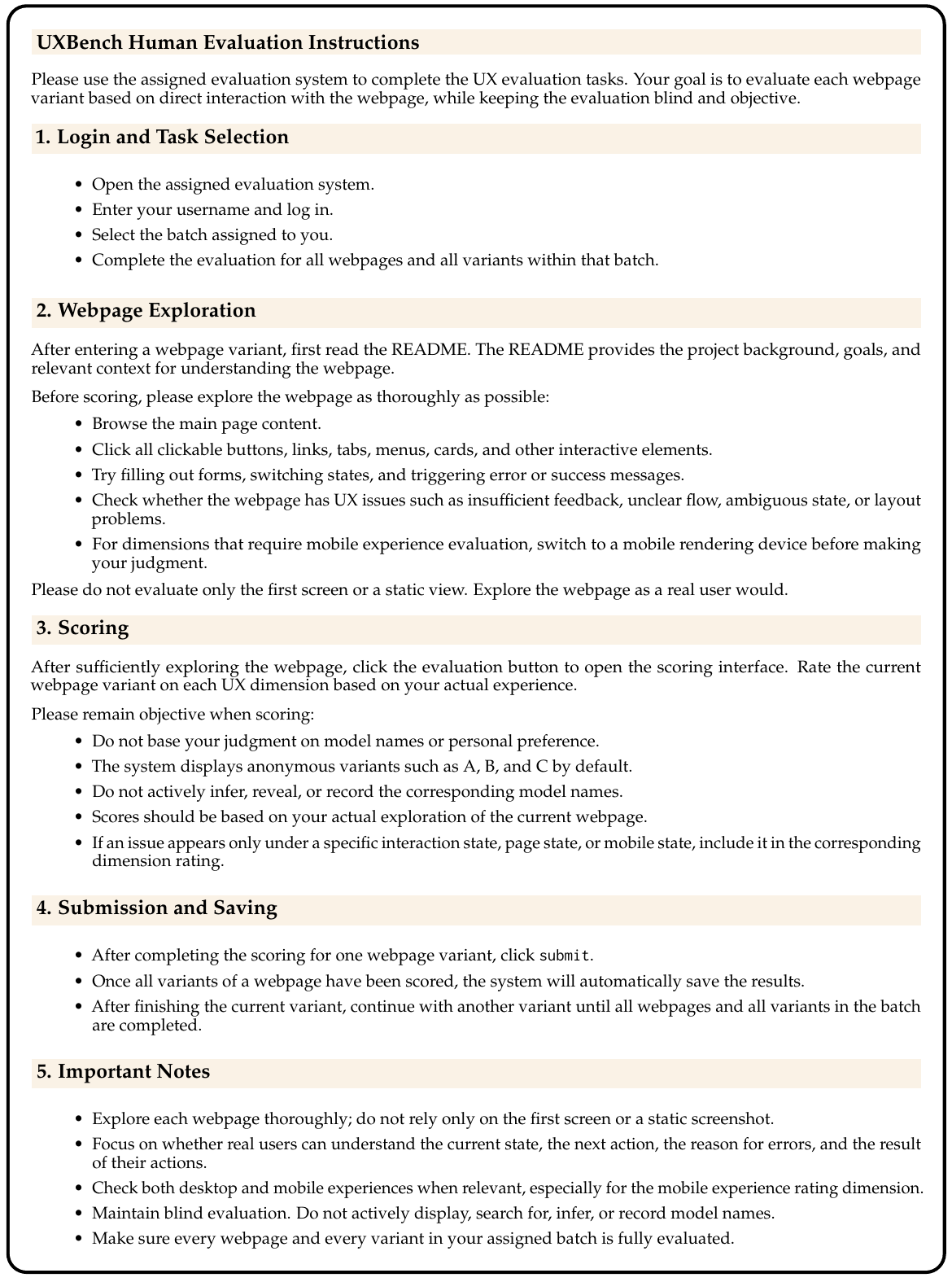}
  \caption{Human Eval Instructions.}
  \label{fig:instruction}
\end{figure*}

\begin{figure*}[t]
  \centering
  \includegraphics[width=\linewidth]{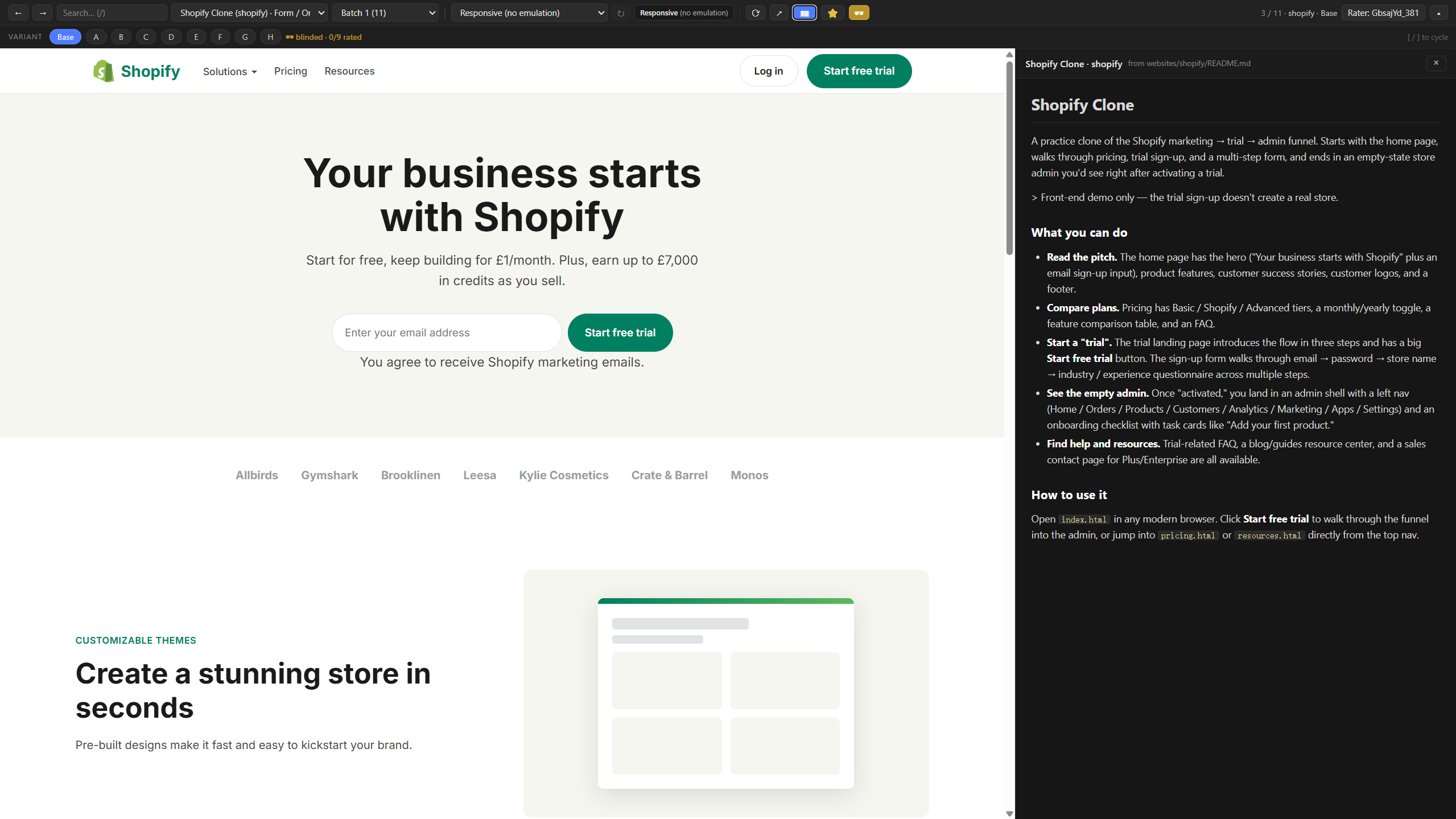}
  \caption{README panel shown alongside each fixture. It describes the project context and intended interactions, helping reviewers understand the use case before evaluating the webpage.}
  \label{fig:readme}
\end{figure*}

\begin{figure*}[t]
  \centering
  \includegraphics[width=\linewidth]{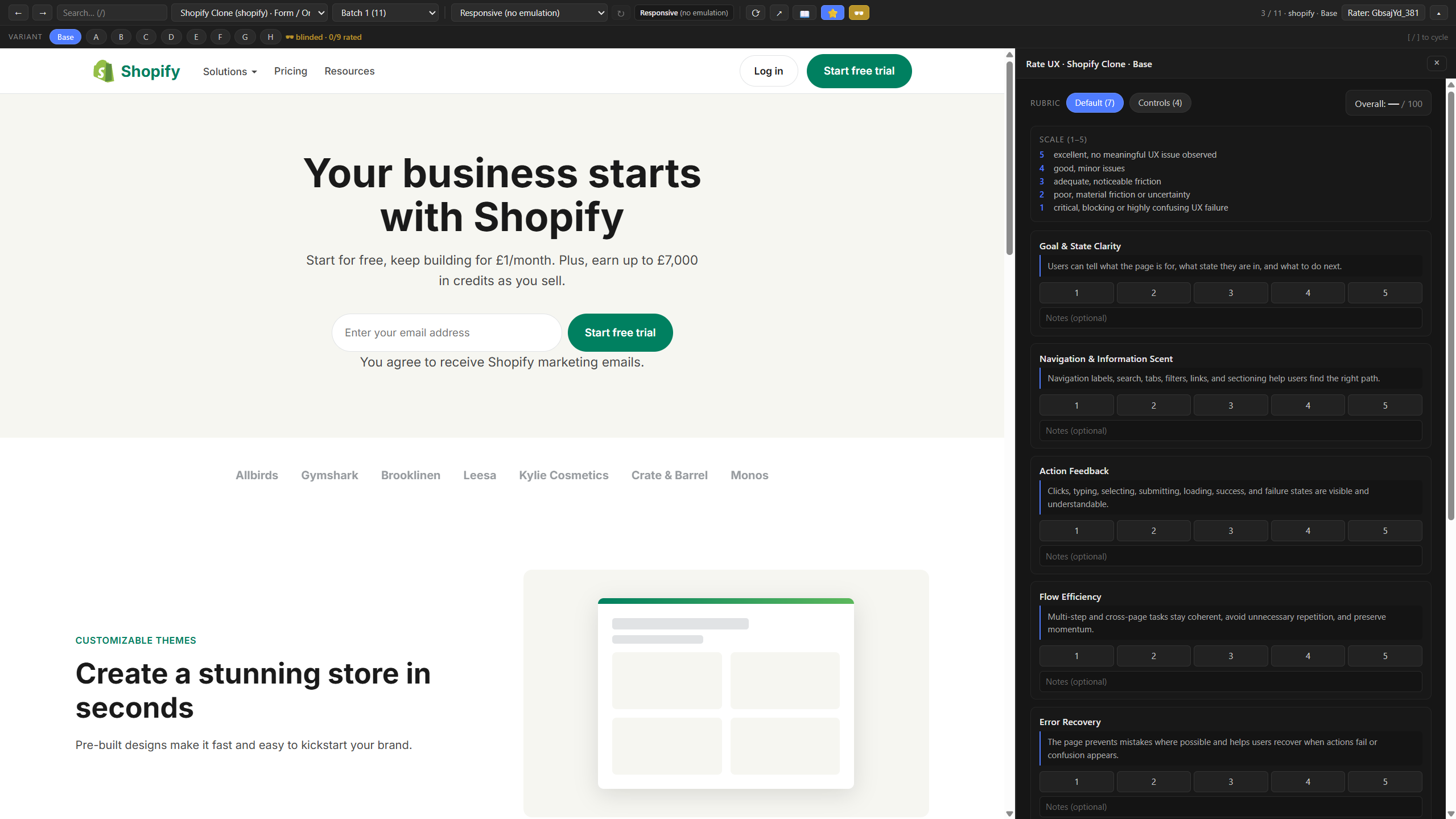}
  \caption{Human rating panel used after exploration. Reviewers score each candidate webpage on the seven UX dimensions using a 1--5 scale, with model identities and automated scores hidden.}
  \label{fig:rating}
\end{figure*}

\begin{figure*}[t]
  \centering
  \includegraphics[width=\linewidth]{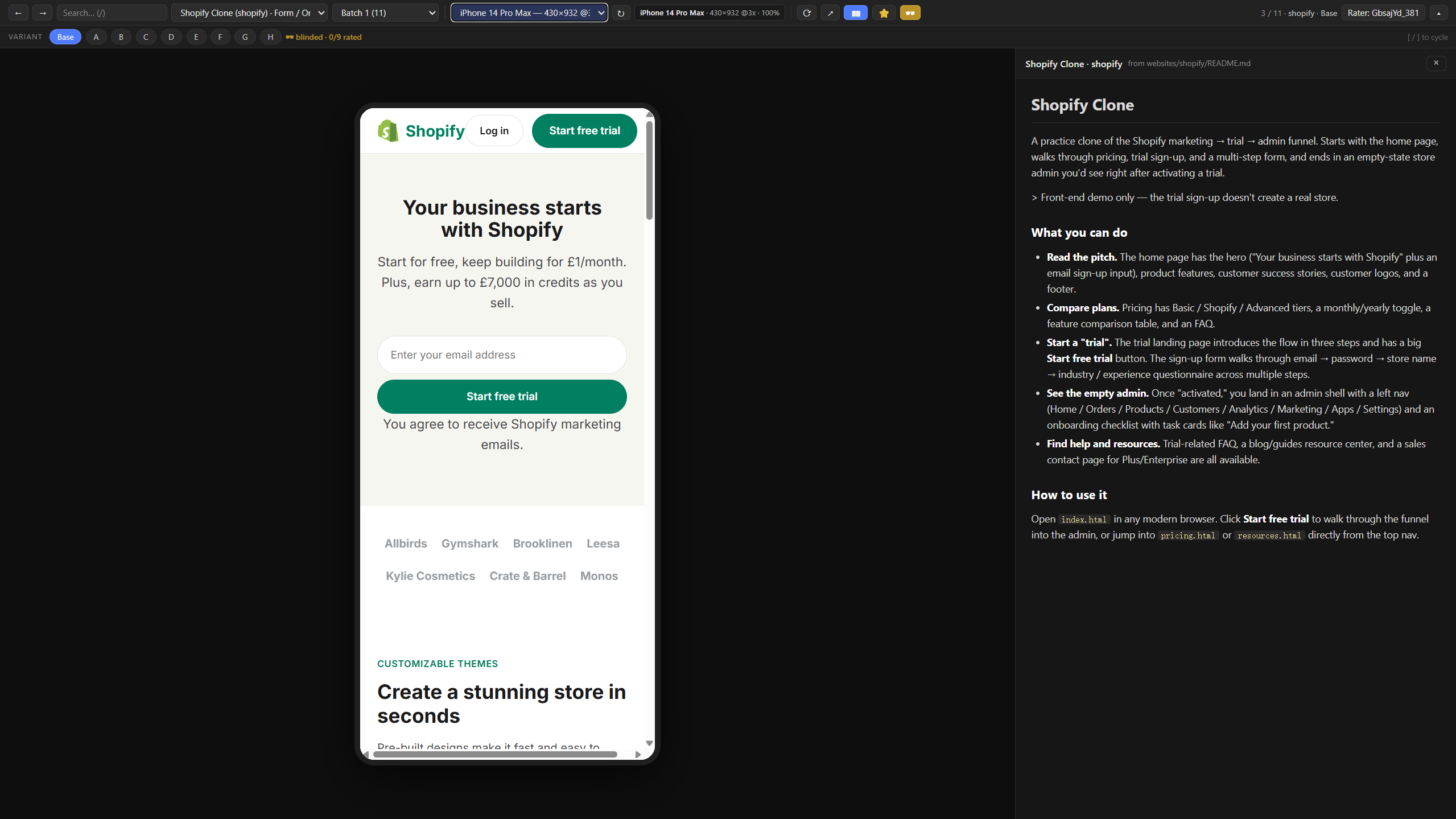}
  \caption{Mobile evaluation view. Reviewers switch the rendering device to inspect responsive layout and mobile-specific interaction behavior before assigning mobile-related ratings.}
  \label{fig:mobile}
\end{figure*}

We provide additional details on the human evaluation interface and procedure. The evaluation instructions is shown in \autoref{fig:instruction}, and interface is shown in \autoref{fig:rating}, \autoref{fig:readme}, \autoref{fig:mobile}. We recruited six participants with UX or front-end design experience. The evaluation was conducted through a custom web interface that presented each webpage together with its README, which described the project context and intended interaction scenario. Participants were asked to read the README before inspecting the webpage, so that their judgments were grounded in the intended use case rather than in visual appearance alone.

Before submitting ratings, participants explored each webpage directly. They were instructed to interact with salient reachable controls, inspect navigation behavior, check visible state changes and feedback messages, and test failure or recovery paths when applicable. After this exploration step, participants opened the scoring panel and rated the candidate webpage on the same seven UX dimensions used in the main evaluation. For dimensions involving mobile experience, participants switched the rendering device in the evaluation interface and evaluated the corresponding mobile layout and interaction behavior.

Model identities were hidden throughout the study. Candidate webpages were shown under anonymous labels such as A, B, and C, and participants could not access model names, automated scores, or aggregate rankings. After all ratings were collected, we restored the hidden label--model mapping and aggregated the human scores across participants, webpages, and rubric dimensions.

\begin{figure*}[t]
    \centering
    \includegraphics[width=0.95\linewidth]{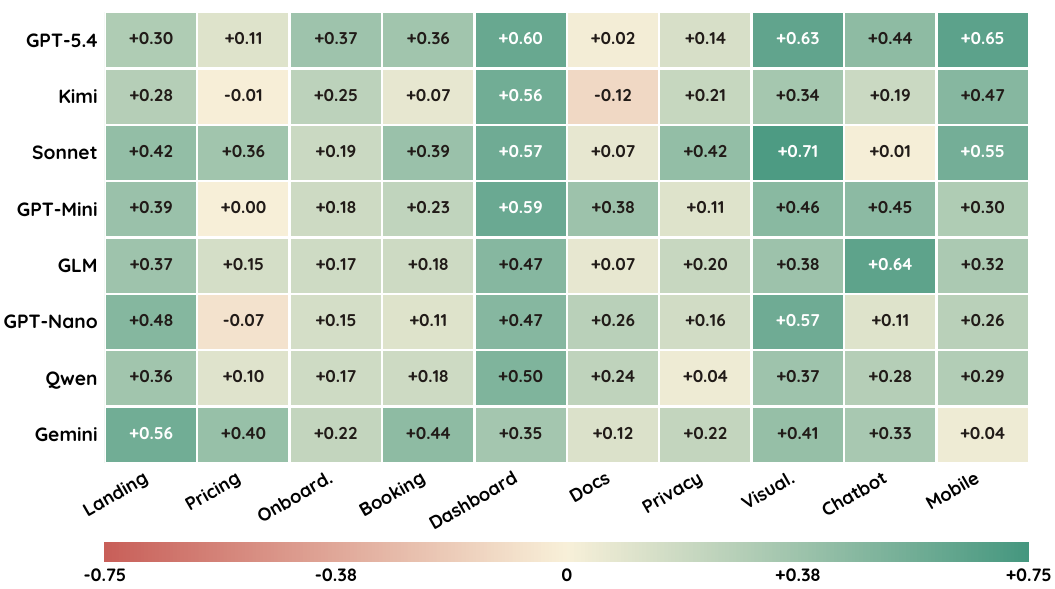}
    \caption{Human--LLM score gaps across model--surface pairs. Each cell shows the difference between blind human ratings and automated LLM-based scores; positive values indicate cases where human reviewers rated the repaired webpage higher than the LLM evaluator.}
    \label{fig:human-llm-delta}
    \vspace{-10pt}
\end{figure*}

\begin{wrapfigure}{r}{0.5\textwidth}
  \centering
  \includegraphics[width=\linewidth]{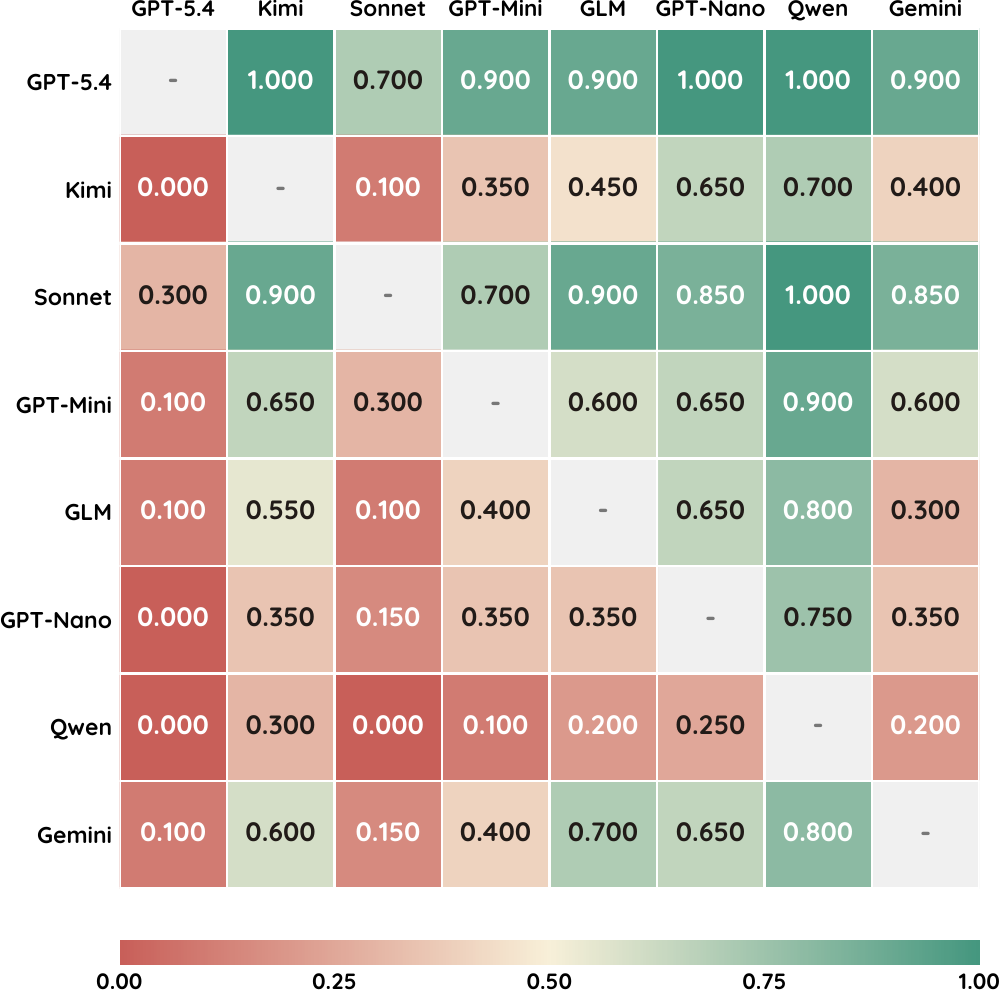}
  \caption{
  Overall human pairwise win-rate matrix.
  }
  \label{fig:human-heatmap}
\end{wrapfigure}

\noindent \textbf{\omark{1} Human pairwise judgments break the automated top-cluster ambiguity.}
The automated LLM-based pairwise results suggest a relatively flat top group, where several strong models remain close in head-to-head comparisons. In contrast, the human pairwise matrix in \autoref{fig:winrate_heatmap} produces a much sharper hierarchy. GPT-5.4 wins nearly all human head-to-head comparisons with large margins, while Claude-Sonnet-4.6 becomes the only model that consistently challenges it. This pattern is also reflected in the human category-level pairwise results in \autoref{tab:human_category_winrate}: GPT-5.4 leads several task- and conversion-oriented surfaces, whereas Claude-Sonnet-4.6 leads multiple information-dense or trust-sensitive surfaces. Thus, human evaluation does not merely reproduce the automated LLM ranking; it separates models whose reports lead to similarly score-improving repairs into those whose repaired webpages are more clearly preferred by expert reviewers.

\noindent \textbf{\omark{2} Human--LLM gaps reveal where automated scoring underestimates perceived UX quality.}
\autoref{fig:human-llm-delta} compares human ratings with automated LLM-based scores at the model--surface level. The differences are not uniform: human ratings are especially higher on visually dense, stateful, or mobile-facing surfaces such as dashboard, visualization, chatbot, and mobile UI, while some simpler surfaces show smaller gaps or even weak reversals for particular models. This indicates that the automated scorer captures a useful repair-actionability signal, but it is less sensitive to some qualities that human reviewers perceive directly in the repaired webpage, such as visual coherence, interaction legibility, and surface-level polish. The comparison therefore calibrates \UXBench rather than contradicting it: automated repair lift is effective for broad model screening, but human evaluation remains necessary for interpreting close ranks and for judging whether repaired interfaces actually feel better to users.

\begin{table*}[t]
  \centering
  \small
  \setlength{\tabcolsep}{6pt}
  \renewcommand{\arraystretch}{1.15}

  \begin{tabularx}{0.98\textwidth}{
    @{}
    >{\raggedright\arraybackslash}p{0.24\textwidth}
    >{\centering\arraybackslash}p{0.16\textwidth}
    >{\centering\arraybackslash}p{0.15\textwidth}
    >{\centering\arraybackslash}p{0.19\textwidth}
    >{\centering\arraybackslash}X
    @{}
  }
    \toprule
    \rowcolor{HeaderGray}
    \textbf{Model}
    & \textbf{Automated Lift}
    & \textbf{Human Mean}
    & \textbf{Automated Rank}
    & \textbf{Human Rank} \\
    \midrule

    \modelcell{GPT-5.4}{openai}
      & +0.216 & 3.84 & 1 & 1 \\

    \rowcolor{RowGray}
    \modelcell{Kimi-K2.5}{kimi}
      & +0.205 & 3.70 & 2 & 6 \\

    \modelcell{Claude-Sonnet-4.6}{claude}
      & +0.174 & 3.81 & 3 & 2 \\

    \rowcolor{RowGray}
    \modelcell{GPT-5.4-Mini}{openai}
      & +0.172 & 3.75 & 4 & 3.5 \\

    \modelcell{GLM-5.1}{glm}
      & +0.171 & 3.75 & 5 & 3.5 \\

    \rowcolor{RowGray}
    \modelcell{GPT-5.4-Nano}{openai}
      & +0.162 & 3.69 & 6 & 7 \\

    \modelcell{Qwen-3.6-Plus}{qwen}
      & +0.148 & 3.67 & 7 & 8 \\

    \rowcolor{RowGray}
    \modelcell{Gemini-3.1-Pro}{gemini}
      & +0.136 & 3.71 & 8 & 5 \\

    \addlinespace[5pt]
    \midrule
    \addlinespace[4pt]
    \multicolumn{5}{@{}l@{}}{
      \footnotesize Spearman's $\rho=0.635$
      \hspace{2.5em}
      Kendall's $\tau_b=0.546$
      \hspace{2.5em}
      $n=8$.
    } \\
    \addlinespace[3pt]
    \bottomrule
  \end{tabularx}

  \caption{
    Agreement between automated repair lift and blind human ratings.
    Automated Lift denotes the model-level repair-score improvement measured automatically.
    Human Mean denotes the average blind human rating.
    Automated Rank and Human Rank denote the corresponding rankings under the two evaluation protocols.
    Rank correlations are computed across the eight model-level aggregates.
  }
  \label{tab:auto_human_agreement}
  \vspace{-10pt}
\end{table*}

\noindent \textbf{\omark{3} Human validation supports automated repair lift as a coarse actionability signal.}
\autoref{tab:auto_human_agreement} compares automated repair lift with blind human ratings at the model-aggregate level. The two signals are positively associated, with Spearman's $\rho=0.635$ and Kendall's $\tau_b=0.546$ across the eight evaluated models, indicating that reports producing larger automated repair gains generally also lead to repaired interfaces that human reviewers perceive as more usable. However, the agreement is not a strict ranking match: GPT-5.4 remains the strongest model under both protocols, but several middle-ranked models shift noticeably between automated and human evaluation. This pattern suggests that automated repair lift captures a useful report-actionability signal, while human ratings capture additional perceptual qualities of the repaired webpage, including clarity, polish, and interaction legibility. Thus, the human study calibrates \UXBench rather than replacing the automated protocol: automated scores are effective for broad model screening, but close model comparisons should be interpreted through human-perceived interface quality.

\section{Statistical Validation of Repair Lift and Human Alignment}
\label{app:stat-validation}

We validate the repair-lift signal with site-paired statistical tests, bootstrap confidence intervals, and model-level human--automatic rank correlations. For both automated and human protocols, lift is computed by pairing each repaired score with the corresponding fixture's unrepaired baseline. This paired design asks whether a judge model's report leads to consistent improvement on the same site, rather than merely achieving a higher unpaired average. Confidence intervals are computed using percentile bootstrap resampling. We report paired $t$-test and Wilcoxon signed-rank $p$-values, together with paired Cohen's $d_z$ as an effect-size measure.

\begin{table*}[t]
  \centering
  \small
  \setlength{\tabcolsep}{5.5pt}
  \renewcommand{\arraystretch}{1.15}

  \begin{tabularx}{0.98\textwidth}{
    @{}
    >{\raggedright\arraybackslash}p{0.235\textwidth}
    >{\centering\arraybackslash}p{0.055\textwidth}
    >{\centering\arraybackslash}p{0.105\textwidth}
    >{\centering\arraybackslash}p{0.16\textwidth}
    >{\centering\arraybackslash}p{0.105\textwidth}
    >{\centering\arraybackslash}p{0.115\textwidth}
    >{\centering\arraybackslash}X
    @{}
  }
    \toprule
    \rowcolor{HeaderGray}
    \textbf{Model}
    & \textbf{$n$}
    & \textbf{Mean Lift}
    & \textbf{95\% boot. CI}
    & \textbf{$t$-test $p$}
    & \textbf{Wilcoxon $p$}
    & \textbf{$d_z$} \\
    \midrule

    \modelcell{GPT-5.4}{openai}
      & 39 & $+0.209$ & $[0.104,\;0.310]$ & $<0.001$ & $<0.001$ & 0.629 \\

    \rowcolor{RowGray}
    \modelcell{Kimi-K2.5}{kimi}
      & 41 & $+0.205$ & $[0.103,\;0.301]$ & $<0.001$ & $<0.001$ & 0.628 \\

    \modelcell{Claude-Sonnet-4.6}{claude}
      & 40 & $+0.188$ & $[0.083,\;0.294]$ & 0.001 & 0.002 & 0.546 \\

    \rowcolor{RowGray}
    \modelcell{GPT-5.4-Mini}{openai}
      & 39 & $+0.188$ & $[0.083,\;0.287]$ & 0.001 & 0.002 & 0.570 \\

    \modelcell{GLM-5.1}{glm}
      & 41 & $+0.171$ & $[0.072,\;0.267]$ & 0.002 & 0.002 & 0.531 \\

    \rowcolor{RowGray}
    \modelcell{GPT-5.4-Nano}{openai}
      & 41 & $+0.162$ & $[0.065,\;0.259]$ & 0.002 & 0.004 & 0.504 \\

    \modelcell{Qwen-3.6-Plus}{qwen}
      & 41 & $+0.148$ & $[0.051,\;0.245]$ & 0.005 & 0.007 & 0.462 \\

    \rowcolor{RowGray}
    \modelcell{Gemini-3.1-Pro}{gemini}
      & 38 & $+0.137$ & $[0.031,\;0.244]$ & 0.018 & 0.027 & 0.400 \\

    \midrule
    \rowcolor{HeaderGray}
    \textbf{Pooled model--site rows}
      & \textbf{320} & $\mathbf{+0.176}$ & $\mathbf{[0.140,\;0.212]}$ & $\mathbf{<0.001}$ & -- & \textbf{0.538} \\

    \bottomrule
  \end{tabularx}

  \caption{Automated repair lift under site-paired comparison against each fixture's unrepaired baseline. Scores are on the 1--5 rubric scale.}
  \label{tab:auto-lift-statistics}
\end{table*}

\begin{table*}[t]
  \centering
  \small
  \setlength{\tabcolsep}{5.5pt}
  \renewcommand{\arraystretch}{1.15}

  \begin{tabularx}{0.98\textwidth}{
    @{}
    >{\raggedright\arraybackslash}p{0.235\textwidth}
    >{\centering\arraybackslash}p{0.055\textwidth}
    >{\centering\arraybackslash}p{0.105\textwidth}
    >{\centering\arraybackslash}p{0.16\textwidth}
    >{\centering\arraybackslash}p{0.105\textwidth}
    >{\centering\arraybackslash}p{0.115\textwidth}
    >{\centering\arraybackslash}X
    @{}
  }
    \toprule
    \rowcolor{HeaderGray}
    \textbf{Model}
    & \textbf{$n$}
    & \textbf{Mean Lift}
    & \textbf{95\% boot. CI}
    & \textbf{$t$-test $p$}
    & \textbf{Wilcoxon $p$}
    & \textbf{$d_z$} \\
    \midrule

    \modelcell{GPT-5.4}{openai}
      & 41 & $+0.346$ & $[0.244,\;0.451]$ & $<0.001$ & $<0.001$ & 1.015 \\

    \rowcolor{RowGray}
    \modelcell{Claude-Sonnet-4.6}{claude}
      & 41 & $+0.318$ & $[0.245,\;0.395]$ & $<0.001$ & $<0.001$ & 1.282 \\

    \modelcell{GPT-5.4-Mini}{openai}
      & 41 & $+0.258$ & $[0.178,\;0.340]$ & $<0.001$ & $<0.001$ & 0.967 \\

    \rowcolor{RowGray}
    \modelcell{GLM-5.1}{glm}
      & 41 & $+0.253$ & $[0.158,\;0.350]$ & $<0.001$ & $<0.001$ & 0.798 \\

    \modelcell{Gemini-3.1-Pro}{gemini}
      & 41 & $+0.214$ & $[0.097,\;0.320]$ & $<0.001$ & $<0.001$ & 0.585 \\

    \rowcolor{RowGray}
    \modelcell{Kimi-K2.5}{kimi}
      & 41 & $+0.211$ & $[0.131,\;0.294]$ & $<0.001$ & $<0.001$ & 0.786 \\

    \modelcell{GPT-5.4-Nano}{openai}
      & 41 & $+0.199$ & $[0.129,\;0.272]$ & $<0.001$ & $<0.001$ & 0.846 \\

    \rowcolor{RowGray}
    \modelcell{Qwen-3.6-Plus}{qwen}
      & 41 & $+0.173$ & $[0.098,\;0.250]$ & $<0.001$ & $<0.001$ & 0.691 \\

    \bottomrule
  \end{tabularx}

  \caption{Human-validation lift under site-paired comparison against each fixture's unrepaired baseline. Scores are blind expert ratings on the same 1--5 rubric scale.}
  \label{tab:human-lift-statistics}
  \vspace{-10pt}
\end{table*}

\begin{wraptable}{r}{0.55\linewidth}
  \centering
  \footnotesize
  \setlength{\tabcolsep}{3pt}
  \renewcommand{\arraystretch}{1.12}

  \begin{tabularx}{\linewidth}{
    @{}
    >{\raggedright\arraybackslash}p{0.34\linewidth}
    >{\centering\arraybackslash}p{0.18\linewidth}
    >{\centering\arraybackslash}p{0.13\linewidth}
    >{\centering\arraybackslash}X
    @{}
  }
    \toprule
    \rowcolor{HeaderGray}
    \textbf{Statistic} & \textbf{Estimate} & \textbf{$p$} & \textbf{95\% boot. CI} \\
    \midrule
    Spearman's $\rho$ & 0.635 & 0.091 & $[-0.190,\;1.000]$ \\
    \rowcolor{RowGray}
    Kendall's $\tau_b$ & 0.546 & 0.061 & $[-0.091,\;1.000]$ \\
    \bottomrule
  \end{tabularx}
  \caption{Model-level rank correlation between automated repair lift and human mean score across the eight evaluated models.}
  \label{tab:human-auto-rank-corr}
\end{wraptable}

\autoref{tab:auto-lift-statistics} and \autoref{tab:human-lift-statistics} show that repair lift is consistently positive under both automated and human protocols. In the automated protocol, the pooled lift across model--site rows is $+0.176$ on the 1--5 scale, with a 95\% bootstrap confidence interval of $[0.140,\;0.212]$ and a paired effect size of $d_z=0.538$. Each individual model also has a positive confidence interval excluding zero. Human validation shows the same direction with larger effects: all repaired interfaces improve over their corresponding site baselines under blind expert ratings, with mean lift ranging from $+0.173$ to $+0.346$ and all paired tests significant. These results support the interpretation that evidence-grounded reports contain actionable information for downstream interface repair.

\autoref{tab:human-auto-rank-corr} compares automated repair lift with human-rated interface quality at the model level. The association is positive but uncertain: Spearman's $\rho=0.635$ and Kendall's $\tau_b=0.546$, but both confidence intervals are wide and include zero. This uncertainty is expected because the rank-correlation analysis contains only eight model-level points. We therefore interpret the automated protocol as a useful actionability signal that is directionally aligned with human judgment, while treating close model rankings as suggestive rather than definitive.



\end{document}